\documentclass{article}

\usepackage{PRIMEarxiv}

\usepackage{amsmath,amsfonts,bm}

\def\eqref#1{equation~\ref{#1}}

\def\1{\bm{1}}

\def\vb{{\bm{b}}}

\def\vw{{\bm{w}}}
\def\vx{{\bm{x}}}

\def\vz{{\bm{z}}}
\def\vF{{\bm{F}}}

\DeclareMathAlphabet{\mathsfit}{\encodingdefault}{\sfdefault}{m}{sl}
\SetMathAlphabet{\mathsfit}{bold}{\encodingdefault}{\sfdefault}{bx}{n}

\usepackage[utf8]{inputenc} 
\usepackage[T1]{fontenc}    
\usepackage{hyperref}       
\usepackage{url}            
\usepackage{booktabs}       
\usepackage{amsfonts}       
\usepackage{nicefrac}       
\usepackage{natbib}
\usepackage{microtype}      
\usepackage{lipsum}
\usepackage{fancyhdr}       
\usepackage{graphicx}       
\usepackage{amssymb}
\usepackage{bbm}
\usepackage{float}
\usepackage{enumitem}
\usepackage{hyperref}
\usepackage{subcaption}
\usepackage{amsmath}
\usepackage{amssymb}
\usepackage{mathtools}
\usepackage{amsthm}
\usepackage{multirow}
\usepackage[normalem]{ulem}
\useunder{\uline}{\ul}{}
\usepackage{arydshln}
\usepackage{arydshln}
\usepackage{algorithm2e}
\RestyleAlgo{ruled}

\title{NeurCAM: Interpretable Neural Clustering via Additive Models}
\author{Nakul Upadhya and Eldan Cohen \\
        \texttt{nakul.upadhya@mail.utoronto.ca}, \texttt{ecohen@mie.utoronto.ca}
        \\University of Toronto, Toronto, Canada}
\begin{document}
\maketitle
\begin{abstract}
Interpretable clustering algorithms aim to group similar data points while explaining the obtained groups to support knowledge discovery and pattern recognition tasks. While most approaches to interpretable clustering construct clusters using decision trees, the interpretability of trees often deteriorates on complex problems where large trees are required. In this work, we introduce the Neural Clustering Additive Model (NeurCAM), a novel approach to the interpretable clustering problem that leverages neural generalized additive models to provide fuzzy cluster membership with additive explanations of the obtained clusters. To promote sparsity in our model's explanations, we introduce selection gates that explicitly limit the number of features and pairwise interactions leveraged. Additionally, we demonstrate the capacity of our model to perform text clustering that considers the contextual representation of the texts while providing explanations for the obtained clusters based on uni- or bi-word terms. Extensive experiments show that NeurCAM achieves performance comparable to black-box methods on tabular datasets while remaining interpretable. Additionally, our approach significantly outperforms other interpretable clustering approaches when clustering on text data.
\end{abstract}

\section{Introduction}

As Machine Learning (ML) has become more prevalent in society in recent years, the need for trustworthy models that stakeholders can audit has increased dramatically. One desirable aspect of trustworthiness in ML is that the approaches utilized are constrained so that their predictive mechanisms are innately understandable to humans. As a result, they are much easier to troubleshoot and more practical for real-world usage \citep{rudin2019stop}. 

One stream of interpretable machine learning is interpretable clustering \citep{yang2021survey}. By using algorithms capable of providing innate explanations of cluster compositions, interpretable clustering methods have found great success in fields such as market segmentation \citep{marketsegmentation}, climate science \citep{sun2024interpretable}, and healthcare \citep{healthcare_use,brain_use}. Most approaches to this task involve the use of decision trees to build clusters \citep{bertsimas2021interpretable,fraiman2013interpretable,frost2020exkmc,gabidolla2022optimal,shati2023optimal,SoftClusterTree}. However, the size of decision trees heavily influences their interpretability \citep{tan2023considerations,luvstrek2016makes}, and complex problems may necessitate larger, less interpretable trees.

Another innately interpretable architecture, the Generalized Additive Model (GAM) \citep{gam_hastie}, has found great success as an interpretable approach in many high-stakes classification and regression tasks \citep{karatekin2019interpretable,sarica2021explainable}. A recent line of work has focused on developing Neural GAMs that enjoy better scalability and are able to learn more expressive, yet interpretable, additive models \citep{NAM,NBM,NODEGAM,ibrahim2024grand}. Despite these benefits, GAMs have not been utilized for clustering.  

In this work, we introduce the \textbf{Neur}al \textbf{C}lustering \textbf{A}dditive \textbf{M}odel (NeurCAM), an interpretable clustering approach that constructs clusters via Neural Generalized Additive Models. Our approach explains how input features influence cluster assignment by modeling the relationship between features and clustering assignments through additive shape functions. Our contributions are as follows: 

\begin{enumerate}
    \item We present a novel approach for interpretable clustering that leverages neural GAMs to provide fuzzy cluster membership. Our approach can leverage deep representations of the data for clustering while still producing explanations in the original feature space. To our knowledge this is the first work to utilize GAMs to provide interpretable clustering.
    \item  We introduce a mechanism that allows users to explicitly constrain the number of single-feature and pairwise interaction shape functions our model utilizes therefore encouraging sparsity in the final explanations, a key quality of interpretable models \citep{rudin2022interpretable}. 
    \item  Through experimentation on a variety of datasets, we demonstrate NeurCAM's effectiveness at creating high-quality clusters when using disentangled representations and also showcase the interpretability provided by additive explanations. 
    \item We demonstrate the capabilities of NeurCAM to perform interpretable text clustering by leveraging transformer-based embeddings in the objective. This allows us to provide uni-word and bi-word explanations while still taking structural and contextual information of the document into account. 
\end{enumerate}

The rest of our paper is organized as follows. In Section \ref{sec:interpretable_clustering} we outline our desiderata for the interpretable clustering task and discuss prior work that aligns with these objectives. In Section \ref{sec:gams} we define GAMs and what makes them interpretable. In Section \ref{sec:neurcam}, we describe the components of our approach, NeurCAM, and in Section \ref{sec:experiments} demonstrate its performance and interpretability. 

\section{Interpretable Clustering}\label{sec:interpretable_clustering}

Our approach for interpretable clustering consists of developing an intrinsically interpretable out-of-sample mapping from samples to clusters. In this section we describe what encompasses this approach, the benefits of achieving it, and discuss existing approaches for interpretable clustering.

\subsection{Interpretable Out-of-Sample Mapping} \label{sec:osm}

Out-of-Sample Mapping (OSM) in clustering refers to the task of assigning a given sample $\vx \in \mathbb{R}^D$ (potentially unseen during training) to a particular cluster using a mapping that is \emph{agnostic of the clustering cost function used} \citep{gabidolla2022optimal}. In particular, OSM allows us to separate the representation of the data used to construct the mapping from samples to clusters from the representation used in the clustering cost function. Previous work has shown that various transformed representations of the original data, such as spectral embeddings, learned embeddings, or PCA, can lead to better clustering performance \citep{spectralclust,subakti2022performance,chang2017feature,SoftClusterTree}. However, such representations are not human-understandable, making it difficult for practitioners to gain insights into the generated clusters. We therefore propose to construct \emph{interpretable} OSM where the mapping is based on interpretable feature representation, while the clustering cost is independently defined over a transformed representation.

\subsection{Model-Based Interpretability} \label{sec:modelbased}

We advocate the use of \emph{intrisically interpretable models} to create the mapping. Formally, our goal is to develop an approach that satisfies the requirements for intrinsic model-based interpretability posed by \citet{murdoch2019definitions}: \emph{modularity, sparsity,} and \emph{simulability}: 
\begin{itemize}
    \item \textbf{Modularity:} A ML model can be considered modular if a user can interpret a meaningful portion of its prediction making process independently from other parts of the network \citep{murdoch2019definitions}.
    \item \textbf{Sparsity:} Sparsity is achieved by limiting the number of non-zero parameters that limit the components a user must analyze to understand model behavior. Understandably, sparse models are easier for practitioners to understand, and hence easier to trust in high-stakes applications \citep{murdoch2019definitions,rudin2022interpretable}.
    \item \textbf{Simulability:} An approach is said to be simulable if a human is able to reasonably internally simulate the entire decision-making process \citep{murdoch2019definitions}. This requirement synergizes with the prior two requirements, as a model often needs to be sparse and modular for a practitioner to be able to recreate its predictions. 
\end{itemize}

In contrast to model-based interpretability, one may opt to use an unintepretable model (e.g. deep neural networks) to assign samples to clusters and apply \emph{post hoc} methods to explain clustering decisions. Although useful in many cases, post hoc approaches face a key problem in practice, where users often have to perform analysis on multiple post hoc explanations to identify which method they should trust \citep{NEURIPS2022_22b11181}. Furthermore, arbitrary post hoc explanations can often be constructed for a given model to obfuscate true biases in the modeling procedure, reducing the trustworthiness of the approach \citep{slack2020fooling}.


\subsection{Existing Approaches}
The most prevalent approach to model-based interpretable clustering involves using unsupervised decision trees to partition the space \citep{bertsimas2021interpretable,fraiman2013interpretable,frost2020exkmc,shati2023optimal,SoftClusterTree,gabidolla2022optimal}. The most relevant approaches to our work are the soft clustering tree (SCT) \citep{SoftClusterTree} and the tree-alternating optimization (TAO) clustering \citep{gabidolla2022optimal}. Both approaches utilize a decision tree to map points to clusters and support OSM with seperate representations. The SCT utilizes an axis-aligned soft-decision tree trained through continuous optimization methods (including mini-batch gradient descent) \citep{SoftClusterTree}, while TAO iteratively refines a tree that approximates a K-means clustering via alternating optimization \citep{gabidolla2022optimal}. 

Despite their interpretable nature, tree-based approaches face a substantial interpretability-performance trade-off as complex problems require large trees to represent cluster boundaries adequately, reducing the sparsity of these approaches. Growing the number of leaves in a tree has been shown to significantly increase the difficulty users face when trying to understand the decision pathways of the model \citep{luvstrek2016makes}. Furthermore, increasing the number of features used in the decision tree significantly increases the time it takes users to analyze the tree and understand what features are essential for predictions \citep{tan2023considerations}. 

Other interpretable clustering approaches include the construction of clusters using polytope machines \citep{lawless2022interpretable} and rectangular rules \citep{chen2016interpretable}. 

Some notable post-hoc approaches include \citet{kauffmann2022clustering} who explains a neural clustering via layerwise relevance propagation \citep{bach2015pixel}, \citet{polyhedral_desc} who profiles assignments via Polyhedral Descriptions, and \citet{carrizosa2022interpreting} who profiles assignments via prototypical examples. Additionally, \citet{deeptextexplain} proposed to cluster text using deep embeddings and then performed a post hoc approximation of the clustering via a logistic regression on a bag-of-words representation. TELL \citep{peng2022xai} and IDC \citep{svirsky2023interpretable} both cluster via a layer in a Neural Network. Similar to post hoc methods, these approaches do not satisfy our desiderata for model-based interpretability (Section \ref{sec:modelbased}).

\section{Generalized Additive Models}\label{sec:gams}
In this work, we propose to construct an interpretable cluster map using a Generalized Additive Model. Given a $D$-dimensional input $\vx = \{x_i\}^D_{i=1}, \vx \in \mathbb{R}^D$, univariate shape functions $f_i$ corresponding to the input features $x_i$, bivariate shape functions $f_{ij}$ corresponding to the features $x_i$ and $x_j$, and link function $g(\cdot): \mathbb{R}^D \rightarrow \mathbb{R}$, the predictions of a GAM and GA$^2$M are defined as follows: 
\begin{gather} 
    \textbf{GAM}: g(\vx) = f_0 + \sum_{i=1}^Df_i(x_i) \\
    \textbf{GA}^2\textbf{M}:g(\vx) = f_0 + \sum_{i=1}^D \left( f_i(x_i) + \sum_{j>i}^D f_{ij}(x_i,x_j) \right)
\end{gather}

In recent years, many powerful GAM models have been proposed, the primary difference between them being how shape functions are constructed. Some notable examples include the Explainable Boosting Machine \citep{EBM}, which uses tree ensembles trained using a cyclical gradient boosting algorithm, and NODE-GAM \citep{NODEGAM}, which leverages layers of ensemble oblivious neural decision trees. 

Additionally, many works have proposed representing the shape function of the GAMs through MLPs such as Neural Additive Models (NAM) \citep{NAM}, which trains an MLP for each feature, and Neural Basis Models \citep{NBM} which extend NAM by constraining all features to utilize a common MLP backbone except for the last layer, which is unique to each feature. 

\paragraph{Interpretability of GAMs}
GAM and GA$^2$Ms satisfy the model-based interpretability requirements previously posed. GAMs force the relationship between the features in the model to be additive, resulting in a \emph{modular} model \citep{murdoch2019definitions}. This modularity allows the contribution of each feature or interaction to the prediction to be visualized as a graph or heatmap, allowing humans to \emph{simulate} how a GAM works by querying the different graphs and adding the results together \citep{lou2013accurate,chang2021interpretable}. This enables decision makers in high-stakes fields such as healthcare to easily understand the explanations provided by GAM and GA$^2$M shape functions \citep{hegselmann2020evaluation,NAM}. Technical stakeholders have also shown a preference for GAMs over post hoc explanations such as SHAP values \citep{shap}, as they reduce the cognitive load needed to grasp a model's decision mechanisms, enhancing stakeholders' confidence in the deployed ML system \citep{kaur2020interpreting}. Furthermore, stakeholders have shown a preference for additive explanations over tree explanations when both utilize a similar number of features \citep{tan2023considerations}. To further ensure that the explanations provided by NeurCAM are \emph{sparse}, we propose selection gates that allow users to enforce a cardinality constraint on the number of features and interactions used.

\section{NeurCAM}\label{sec:neurcam}
\begin{figure*}[!ht]
    \centering
    \includegraphics[width = \textwidth]{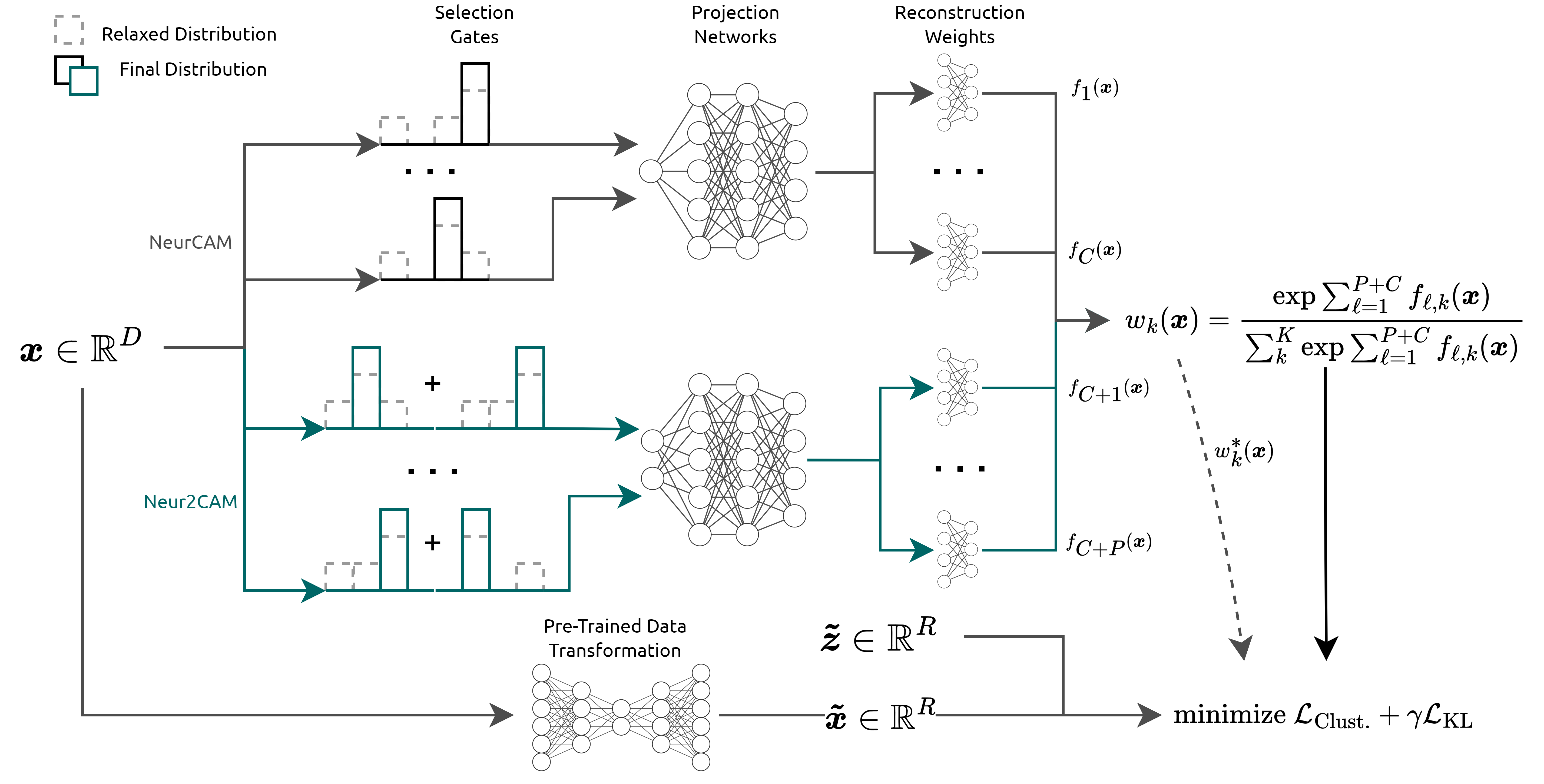}
    \caption{Our proposed approach. We leverage multiple shape functions that each pick a feature via a selection gate. The final prediction is the sum of the individual shape function contributions. The black shape functions represents NeurCAM and the additional blue pairwise shape functions represent Neur2CAM.}
    \label{fig:NeurCAM}
\end{figure*}
In this work, we consider the following problem: Let $X = \{\vx_n\}_{n=1}^N$ be a set of $N$ data points with $\vx_n$ being a $D$-dimensional feature vector $\vx_n \in \mathbb{R}^D$. We aim to decide fuzzy cluster assignments $w_{n,k} \in [0,1],\ \sum_k w_{n,k} = 1.0$ for each point $\vx_n$ and cluster $k \in 1,2,\ldots, K$. Our approach to this problem involves utilizing a Neural GAM to obtain the assignments (Section \ref{sec:GAMdesign}) and train this GAM through a combination of a fuzzy clustering loss and self-supervised regularization (Section \ref{sec:training}). 

\subsection{Model Architecture}\label{sec:GAMdesign}
In the following section, we describe how we obtain the cluster assignments via an interpretable neural GAM. For succinctness, we drop the sample index $n$ in this section. 
\subsubsection{Neural Basis Model}
For our additive model, we leverage a Neural Basis Model (NBM) \citep{NBM} which we describe here for thoroughness. 

The NBM operates by projecting each feature $x_i, i=1,2,\ldots,D$ onto $B$ basis functions that are shared between all features. The projection function $\vb(\cdot):\mathbb{R} \rightarrow \mathbb{R}^B$ is represented by an MLP backbone with a single input and $B$ outputs. The projection of each feature is then reconstructed into the final shape function for each cluster via a linear combination, with each feature having its own set of weights for each cluster $\bm{\lambda}_{i,k} \in \mathbb{R}^{B}$. More concretely, the prediction from a single feature $x_i$ is as follows:
\begin{gather}
    \vb(x_i) = \text{MLP}(x_i) \\
    f_{i,k}(\vx) = \bm{\lambda}_{i,k}\cdot \vb(x_i)
\end{gather}

The resultant logits assigned to each cluster $h_k(\vx), k = 1,2,\ldots,K$ is the sum of the contributions from each feature: 
\begin{gather} 
    g_k(\vx) = \sum_{i}^D f_{i,k}(\vx)
\end{gather}
These logits are transformed via the softmax operation to obtain the final fuzzy assignment weights: 

\begin{gather}\label{eqn:softmax}
    w_{k}(\vx_n) = \frac{\exp \left( g_k(\vx) \right)}{\sum_{k'=1}^K \exp \left(g_{k'}(\vx) \right)}
\end{gather}
Pairwise interaction can be represented in a similar manner using an MLP with two inputs instead of one. 

\subsubsection{Feature Selection Gates} \label{sec:gates}

A common problem across MLP-based GAMs and GA$^2$Ms is the variable explosion problem. As the dimensionality of the dataset increases, the number of single-feature and pairwise shape functions rapidly increases as well. This is especially true for GA$^2$Ms as the number of pairwise interactions grows quadratically with the dimensionality of the dataset. Previous MLP-based approaches have no inherent mechanisms to limit pairwise interactions and instead utilize heuristics based on residual analysis of a fitted GAM to select interactions to use in the GA$^2$M \citep{NAM,NBM}. However, this approach does not translate to the unsupervised setting as there are no residuals due to the absence of true labels. 

Instead, we propose to \emph{learn} what features and interactions are in our model through feature selection gates. More concretely, NeurCAM learns $C$ independent shape functions. The feature used by shape function $c \in 1,\ldots,C$ is selected via a selection function $s_c(\vx): \mathbb{R}^D \rightarrow \mathbb{R}$. A common way to represent such a function is the product of the feature vector and a one-hot selection vector $\vF_c$ where a variable is selected if its corresponding selection logit $\Tilde{F_c} \in \mathbb{R}^D$ is the largest:
\begin{gather}\label{eqn:onehot}
    s_c(\vx) = \vF_c \cdot \vx \\ 
    F_{c,i} = \begin{cases}
        1 & i = \text{argmax}(\Tilde{\vF_c}) \\
        0 & \text{else}
    \end{cases}
\end{gather}
To train this selection as part of our network, we \emph{temporarily} relax the one-hot vector to be represented by entmax$_{\alpha}$ \citep{entmax}, a sparse version of softmax that allows elements to become exactly zero if the logits are sufficiently small: 
\begin{gather} \label{eqn:sel1}
    s_c(\bm{x}) = \text{entmax}_{\alpha}(\Tilde{\vF_c} / T) \cdot \vx 
\end{gather}
Here, $T > 0$ is a temperature annealing parameter that controls the sparsity of the distributions obtained from the entmax$_\alpha$ operations, with smaller values of $T$ resulting in sparser distributions. As $T \rightarrow 0$, the resultant distribution will become one-hot. 

At the start of training, we set $T=1.0$, making the gate a weighted mixture of features. After a number of warm-up training epochs, we anneal $T$ by a factor of $\epsilon \in (0,1)$ until $\text{entmax}_{\alpha}(\Tilde{\vF_c} / T)\cdot\vx = x_{\text{argmax}(\Tilde{\vF_c})}$, allowing $s_c(\cdot)$ to serve as a proper feature selection gate and making NeurCAM a valid GAM once again. The factor $\epsilon$ is a hyperparameter that controls how gradual the tempering is, with higher epsilon values resulting in a slower annealing process and lower values resulting in a faster annealing. 

We modify the NBM to include these selection gates to obtain the additive model architecture used by NeurCAM whos prediction mechanisms are as follows:
\begin{gather}
    \vb(\vx) = \text{MLP}(s_c(\vx)) \\
    f_{c,k}(\vx) = \bm{\lambda}_{c,k}\cdot \vb(\vx) \\
    g_k(\vx) = \sum_{c=1}^Cf_{c,k}(\vx)
\end{gather}
A softmax is then applied to $g_1(\vx),g_2(\vx)\ldots,g_k(\vx)$ to obtain the soft cluster assignments $\vw(\vx)$ like in Equation (\ref{eqn:softmax}).

The number of shape functions serves as an upper bound for the number of features utilized by the model, and the features used in the model's explanations can be limited by setting $C < D$. The choice of $C$ is problem and stakeholder specific. 

While we apply this mechanism on an NBM, these selection gates can be applied to other Neural GAM models such as the Neural Additive Model \citep{NAM}. 

\paragraph{Extending to Neur2CAM:}
To extend NeurCAM to allow pairwise interactions, we introduce $P$ additional shape functions whose selection gates allow for two features. More concretely, the pairwise interaction used by shape function $p \in C+1, \ldots, C+P$ is chosen by selection function $s_p^2(\vx): \mathbb{R}^D \rightarrow \mathbb{R}^2$, defined as: 
\begin{gather} \label{eqn:sel2}
    s_p^2(\vx) = \begin{bmatrix}
    \text{entmax}_\alpha(\Tilde{\vF}_{p,0} / T_2) \\ 
    \text{entmax}_\alpha(\Tilde{\vF}_{p,1} / T_2)
    \end{bmatrix}\vx
\end{gather}
Like the single-order case, the pairwise temperature annealing parameter $T_2$ is set to 1.0 during the warmup phase and is annealed to near zero afterwards. All pairwise shape function share a common two-input MLP backbone that projects a pair of features into $B$ outputs, with each shape function having its own set of reconstruction weights used to build the final shape functions. In theory, our approach can extend NeurCAM to interactions of any order, but we only explore up to pairwise to maintain a high degree of interpretability. 

\subsection{Training NeurCAM}\label{sec:training}
In this section we describe the loss function and procedure used to train NeurCAM. 
\subsubsection{Fuzzy Clustering Loss}
We employ a loss function inspired by Fuzzy C-Means \citep{fuzzycmeans}:

\begin{gather}\label{eqn:loss}
    \mathcal{L}_{\text{Clust}} =  \sum_{n=1}^{N} \sum_{k=1}^{K}w_{k}(\vx_n)^m ||\vx_n - \vz_k||^2 
\end{gather}

Here $m \geq 1.0$ is a hyperparameter controlling the fuzziness of the clustering. Our loss departs from Fuzzy C-Means in two main ways. In constrast to Fuzzy C-Means where the cluster assignment weights are free variables, our assignment weights are parameterized by an interpretable GAM (Equation (\ref{eqn:softmax})). 

Additionally, instead of the centroids of the clusters being calculated by the clustering assignments \citep{fuzzycmeans}, we opt to make the centroids $\vz_k \in \mathbb{R}^D, k = 1,\ldots, K$ free variables to aid in the optimization of our network.

\subsubsection{Disentangling Representations} 
The mapping constructed by NeurCAM is agnostic to how the distance between samples and the cluster centroids are defined. As such, we decouple the representations of the data in our loss function into the interpretable representation $\vx$ and the transformed representation $\Tilde{\vx} \in \mathbb{R}^R$. 

NeurCAM maps a sample to a given cluster using the interpretable feature set. However, the distance from each sample to the centroids of the clusters will be calculated using the transformed representation:
\begin{gather}\label{eqn:decouploss}
    \mathcal{L}_{\text{Clust}} = \sum_{n=1}^{N} \sum_{k=1}^{K}w_{k}(\vx_n)^m ||\Tilde{\vx}_n - \Tilde{\vz}_k||^2
\end{gather}
It is important to note that the learned centroids $\Tilde{\vz}_k \in \mathbb{R}^R$ are in the same space as the transformed representation $\Tilde{\vx}$. 

\subsubsection{Self-Supervised Regularization}
Empirically, we observe that while our model is able to achieve a high-quality clustering at the end of the warm-up period, the selection gate annealing process often significantly degrades the ability of our network to directly optimize the clustering loss (Equation (\ref{eqn:decouploss})) and can lead to poor local optima. To mitigate this degradation, we propose to take advantage of the clustering discovered at the end of the warm-up period, denoted by $\vw^*(\vx_n)\in \mathbb{R}^K$, to guide the optimization process after the annealing process starts. 

When we start annealing the selection gates' temperatures $T$ and $T_2$ toward zero, we add a regularization term that penalizes the KL-divergence between the current mapping $\vw(\vx_n)$ and the mapping $\vw^*(\vx_n)$ discovered at the end of the warm-up phase. 

\begin{gather} \label{eqn:kl}
    \mathcal{L}_{\text{KL}} = \sum_{n=1}^{N} \sum_{k=1}^{K}w_{k}^*(\vx_n)\log \frac{w_{k}^*(\vx_n)}{w_{k}(\vx_n)}
\end{gather}
The final objective utilized after the warmup phase is as follows: 
\begin{gather}\label{eqn:secondphase}
\text{minimize}\  \mathcal{L}_{\text{Clust}} + \gamma\mathcal{L}_{\text{KL}}
\end{gather}
Where $\gamma$ is a parameter that controls the weight of the KL term. The complete pseudocode of NeurCAM's training procedure can be found in Algorithm (\ref{alg:gate_annealing}).
\begin{algorithm}
\caption{NeurCAM Training Pseudocode}\label{alg:gate_annealing}
\KwIn{$X, \Tilde{X}, K, \gamma,\alpha, \epsilon, E_\text{warmup}, E_\text{total}$}
$\Tilde{\vz} \gets \text{InitializeCentroids}(\Tilde{X})$\\
$\theta_\text{GAM} \gets \text{RandomInitModelParams()}$\\
$\theta \gets \Tilde{\vz} \cup \theta_\text{GAM}$ \\
$T \gets 1.0$\\
\For{$E = 1, 2,\ldots,E_{\text{total}}$}{
    \If{$E = E_\text{warmup}$}{
        $\theta^* \gets \theta$
    }
    \For{$(X_{\text{batch}}, \Tilde{X}_\text{batch}) \in (X, \Tilde{X})$ }{
    
        $\vw_\text{batch} \gets $ ForwardPass$(X_{\text{batch}}, \theta, T)$\\

        \eIf{$E > E_\text{warmup}$}{
            $\vw^*_\text{batch} \gets$ ForwardPass$(X_\text{batch}, \theta^*, 1.0)$
            $\mathcal{L}_{\text{KL}} \gets \text{KL}(\vw^*_\text{batch}||\vw_\text{batch})$ \\
        }{
            $\mathcal{L}_{\text{KL}} \gets 0$
        }
        $\mathcal{L}_\text{Clust} \gets \text{CalculateClustLoss}(\Tilde{X}_{\text{batch}}, \vw_{\text{batch}}, \Tilde{\vz})$\\
        $\mathcal{L} \gets \mathcal{L}_\text{clust} + \gamma\mathcal{L}_{\text{KL}}$\\
        $\theta \gets \theta - \alpha \nabla \mathcal{L}$
    }
                
    \If{$\exists c\ s_c(\vx) \neq \vx_{\text{argmax}(\Tilde{\vF_c})}$}{$T \gets \epsilon T$}
    
}
\end{algorithm}
\section{Experiments and Evaluation}\label{sec:experiments}
In this section, we highlight the benefits of our approach in various real-world datasets. We first demonstrate our ability to generate high-quality clusters using disentangled representations on tabular data sets. We later extend our approach to text clustering. In addition, we validate our training scheme via an ablation study and provide an analysis of the interpretability of NeurCAM.

\subsection{Experimental Details} 
We run experiments with two variants of our approach. NeurCAM (NCAM) includes only single-feature shape functions, and Neur2CAM (N2CAM), which extends NeurCAM to include pairwise interaction.  

For our interpretable benchmarks, we consider approaches that provide model-based interpretability and are capable of disentangling representations, namely the recently proposed Soft Clustering Trees (SCT) \citep{SoftClusterTree} and the axis-aligned TAO Clustering Tree (TAO) \citep{gabidolla2022optimal}. For both of these tree-based approaches, we consider two different depths. We first consider a highly interpretable shallow tree with a depth of five (SCT/TAO-5). This choice also guarantees that the number of leaves is greater than the number of clusters across all datasets. For a more expressive, but less sparse, baseline, we also consider trees with a depth of seven (SCT/TAO-7). As a representative of black-box clustering methods, we compare with Mini-Batch $K$ -Means (mKMC) \citep{sculley2010web}, a scalable variant of K-Means.

\paragraph{Evaluation Metrics}
As all of our utilized datasets come with known labels, we assess the clustering results using three external evaluation measures: Adjusted Rand Index (ARI), Normalized Mutual Information (NMI), and Unsupervised Clustering Accuracy (ACC). We also report the Inertia (Iner.) normalized by the number of datapoints in the dataset. For our fuzzy models (NeurCAM and SCT), we calculate all metrics using the hard clustering decision obtained by selecting the cluster with the highest fuzzy weight for each data point at inference time.

Rand index (RI) \citep{RI} measures agreements between two partitions of the same dataset $P_1$ and $P_2$ with each partition representing $\binom{n}{2}$ decisions over all pairs, assigning them to the same or different clusters and is defined as follows:
\begin{gather}
    \text{RI}(P_1, P_2) = \frac{a+b}{\binom{n}{2}}
\end{gather}
Where $a$ is the number of pairs assigned to the same cluster, and $b$ is the number of pairs assigned to different  clusters. ARI \citep{ARI} is a correction for RI based on its expected value: 
\begin{gather}
    \text{ARI} = \frac{\text{RI} - \mathbb{E}(\text{RI})}{\max(\text{RI}) - \mathbb{E}(\text{RI})}
\end{gather}
An ARI score of zero indicates that the cluster assignment is no better than a random assignment, while a score of 1 indicates a perfect match between the two partitions. 

Normalized Mutual Information measures the statistical information shared between distributions \citep{nmi}, normalized by the average entropy of the two distributions. This metric is defined as follows: 
\begin{gather}
    \text{NMI}(P_1, P_2) = \frac{\text{MI}(P_1, P_2)}{\text{mean}(H(P_1),H(P_2))}
\end{gather}
Where $H(\cdot)$ is the entropy of a given distribution and $\text{MI}(P_1, P_2)$ is the mutual information between $P_1$ and $P_2$.

Unsupervised clustering accuracy \citep{xie2016unsupervised} measures the best agreement between the cluster label $c_n$ and the ground-truth $l_n$.
\begin{gather}
    \text{ACC} = \underset{map \in M}{\max} \frac{1}{N}\sum_{n=1}^N\mathbbm{1} \{l_n = map(c_n)\} 
\end{gather}
$M$ is the set of all possible one-to-one mappings from clusters to ground-truth labels. 

Inertia is defined as the sum of the squared distances from the representation of the datapoints and the centroids of the cluster they are assigned to:
\begin{gather}
    \text{Inertia} = \sum_{n=1}^N\sum_{k=1}^K w_{n,k} ||\Tilde{\vx}_n - \vz_k||\\
    w_{n,k} \in \{0,1\},\ \sum_{k=1}^K w_{n,k} = 1\ \forall n = 1,2,\ldots,N
\end{gather}

\paragraph{Training  and Implementation Details}
All models and benchmarks are implemented in Python. The SCT and NeurCAM are implemented in PyTorch \citep{Pytorch} and we utilize the Mini-Batch $K$-Means implementation found in the Scikit-Learn package \citep{scikit-learn}. For TAO, the implementation from the original paper is not publicly available and we implemented the approach following the details outlined by \citet{gabidolla2022optimal}. 

NeurCAM is trained using the Adam \citep{adam} optimizer with plateau learning rate decay. For Neur2CAM, the pairwise gate temperature parameter $T_2$ is fully annealed before $T$ starts its annealing procedure. To initialize the centroids $\Tilde{\vz}$, we utilize centroids obtained from Mini-Batch $K$-Means clustering. Training details and hyperparameters for NeurCAM and the other approaches can be found in Appendix \ref{sec:hp}.

As our model and all benchmarks may converge to a locally optimum solution, we perform five runs with different random seeds and select the run with the lowest Inertia value for the hard clustering. 

\paragraph{Extracting Shape Graphs:} To extract the final shape graphs for each feature and interaction in NeurCAM, we query the predictions from each inidividual shape function and then combine the shape functions that have selected the same feature (or interaction) to obtain the final shape graphs: 
\begin{gather}
    f_{i,k}(\vx) = \sum_{c=1}^C\mathbbm{1}_i(\vF_c)f_{c,k}(\vx) \\
    f_{i,j,k}(\vx)  = \sum_{p=C+1}^{C+P}\mathbbm{1}_{i,j}(\vF_{p,0}, \vF_{p,1})f_{p,k}(\vx)
\end{gather}
Here $\mathbbm{1}_i(\cdot)$ is an indicator function on whether feature $i$ is selected and $\mathbbm{1}_{i,j}(\cdot,\cdot)$ is an indicator function on whether both features $i$ and $j$ were selected in the selection vectors (regardless of the order). Following \citet{NAM} and \citet{NBM}, we set the average cluster activation (logits) of each feature's shape function to zero by subtracting the mean activation. For the pairwise shape graphs of Neur2CAM, we adopt GA$^2$M purification to push interaction effects into main effects if possible \citep{purify}. To derive feature-importance from our model, we follow \citet{EBM} and take the average absolute area under the shape graph.

\subsection{Clustering Tabular Data}

We demonstrate the ability of our model to create high-quality cluster assignments on tabular tasks by testing it on six datasets from the UCI repository \citep{UCI}: Adult, Avila, Gas Drift, Letters, Pendigits, and Shuttle. All datasets were standardized by removing the mean and scaling to unit variance. Information about these datasets can be found in Appendix \ref{sec:data_info_appendix}. 

In this set of experiments we set the number of single-feature shape functionss equal to the number of features. For Neur2CAM, we set the number of pairwise shape functions equal to the number of single-feature shape functions to maintain a high degree of interpretability. We set $m=1.05$ for our loss function (Equation (\ref{eqn:decouploss})).

\paragraph{Representations Utilized:} For our interpretable representation $\vx$, we utilize the original feature space provided by the datasets. For the representation in our loss function $\Tilde{\vx}$, we consider two different deep transformations:  
\begin{itemize}
    \item Denoising AutoEncoder (DAE): We cluster on the embeddings from a pre-trained DAE \citep{vincent2008extracting} with dropout corruption and a bottleneck of size 8. 
    \item SpectralNet (Spectral): We cluster on embeddings from a SpectralNet \citep{spectralnet}, a deep-learning based approximation of Spectral Clustering \citep{spectralclust}. The embedding dimension is equal to the number of clusters in the dataset.
\end{itemize}

NeurCAM, the SCT, and TAO make clustering decisions using the original, interpretable feature space, while Mini-Batch $K$-Means directly clusters in the transformed space.

\subsubsection{Results}
\begin{table*}[!ht]
\centering
\small
\def\arraystretch{.95}
\caption{Summarized results from our tabular clustering experiments. The best overall results are highlighted in bold, while the best interpretable results are underlined. The dashed line separates interpretable and non-interpretable approaches.}
\begin{tabular}{@{}cccccccccl@{}}
\toprule
                                &       & \multicolumn{2}{c}{ARI} $\uparrow$     & \multicolumn{2}{c}{NMI} $\uparrow$     & \multicolumn{2}{c}{ACC} $\uparrow$     & \multicolumn{2}{c}{Iner.}  $\downarrow$                         \\ \midrule
                                &       & AE    & Spectral             & AE    & Spectral             & AE    & Spectral             & AE                   & \multicolumn{1}{c}{Spectral} \\ \midrule
\multirow{7}{*}{Average Scores} & NCAM  & 0.192 & 0.261                & 0.303 & 0.385                & 0.524 & 0.561                & 2.509                & 1.995                        \\
                                & N2CAM & 0.190 & {\ul \textbf{0.272}} & 0.295 & {\ul \textbf{0.394}} & 0.517 & {\ul \textbf{0.568}} & 2.399                & {\ul 1.994}                  \\
                                & TAO-5 & 0.185 & 0.240                & 0.287 & 0.347                & 0.521 & 0.515                & 2.960                & 4.181                        \\
                                & TAO-7 & 0.189 & 0.257                & 0.291 & 0.371                & 0.526 & 0.518                & {\ul \textbf{2.315}} & 2.517                        \\
                                & SCT-5 & 0.159 & 0.190                & 0.276 & 0.314                & 0.452 & 0.479                & 3.552                & 5.460                        \\
                                & SCT-7 & 0.157 & 0.198                & 0.267 & 0.315                & 0.469 & 0.476                & 3.416                & 4.443                        \\\cdashline{2-10} \noalign{\smallskip}
                                & mKMC  & 0.211 & 0.250                & 0.301 & 0.380                & 0.512 & 0.520                & 2.682                & \textbf{1.579}               \\ \midrule
\multirow{7}{*}{Average Rank}   & NCAM  & 9.50  & 4.17                 & 8.17  & 3.33                 & 7.00  & 4.67                 & 2.83                 & {\ul 3.17}                   \\
                                & N2CAM & 8.83  & {\ul \textbf{3.00}}  & 7.83  & {\ul \textbf{3.00}}  & 7.67  & {\ul \textbf{3.50}}  & {\ul 2.67}           & 3.50                         \\
                                & TAO-5 & 9.67  & 6.50                 & 9.50  & 5.83                 & 6.83  & 7.83                 & 4.67                 & 4.50                         \\
                                & TAO-7 & 9.50  & 5.67                 & 9.50  & 4.33                 & 7.00  & 7.00                 & 3.00                 & {\ul 3.17}                   \\
                                & SCT-5 & 9.83  & 6.00                 & 10.83 & 6.33                 & 9.83  & 9.50                 & 6.50                 & 6.50                         \\
                                & SCT-7 & 11.33 & 8.50                 & 12.00 & 7.33                 & 10.17 & 9.67                 & 5.83                 & 5.33                         \\\cdashline{2-10} \noalign{\smallskip}
                                & mKMC  & 7.00  & 4.00                 & 8.17  & 3.33                 & 7.33  & 6.00                 & \textbf{2.33}        & \textbf{1.67}                \\ \bottomrule
\end{tabular}
\label{tab:sum_tab_cluster}
\end{table*}

Table \ref{tab:sum_tab_cluster} provides a summary of the comparison between our approach and our baselines on the tabular data sets. A detailed breakdown of the results by dataset can be found in Appendix \ref{sec:tab_results}.

We observe that our approaches on SpectralNet embeddings consistently outperforms baselines in average rank on external metrics across all datasets. This trend remains consistent when looking at the average value across data sets, as well. 

When considering Inertia for SpectralNet embeddings, we achieve the best average value and average rank across the interpretable methods. Although TAO-7 is able to obtain a slightly lower Inertia value on DAE embeddings, we still achieve a lower average rank, indicating that our approaches result in lower Inertia more often.

\subsection{Clustering Text Data}

\begin{table}[!ht]
\centering
\small
\def\arraystretch{.95}
\caption{Summarized results from our text clustering experiments. The best overall results are highlighted in bold, while the best interpretable results are underlined. The dashed line separates interpretable and non-interpretable approaches.}
\begin{tabular}{@{}cccccc@{}}
\toprule
\multicolumn{1}{l}{}            & \multicolumn{1}{l}{} & ARI $\uparrow$          & NMI $\uparrow$               & ACC $\uparrow$           & Iner. $\downarrow$                   \\ \midrule
\multirow{8}{*}{Average Values} & NCAM                 & 0.359          & 0.463               & 0.584          & 696.717                 \\
                                & N2CAM                & {\ul 0.455}    & {\ul 0.546}         & {\ul 0.625}    & {\ul 686.311}           \\
                                & TAO-5                & 0.052          & 0.236               & 0.286          & 737.509                 \\
                                & TAO-7                & 0.081          & 0.272               & 0.346          & 730.358                 \\
                                & SCT-5                & 0.127          & 0.233               & 0.333          & 738.614                 \\
                                & SCT-7                & 0.134          & 0.224               & 0.320          & 737.754                 \\ \cdashline{2-6} \noalign{\smallskip}
                                & mKMC                 & \textbf{0.491} & \textbf{0.570}      & \textbf{0.659} & \textbf{674.071}        \\\midrule
\multirow{7}{*}{Average Rank}   & NCAM                 & 3.00           & 3.00                & 2.75           & 3.0                     \\
                                & N2CAM                & {\ul 1.75}     & {\ul \textbf{1.50}} & {\ul 2.00}     & {\ul 2.0}               \\
                                & TAO-5                & 6.25           & 6.00                & 6.00           & 6.0                     \\
                                & TAO-7                & 5.25           & 4.75                & 4.75           & 4.5                     \\
                                & SCT-5                & 5.50           & 5.00                & 5.50           & 6.0                     \\
                                & SCT-7                & 5.00           & 6.25                & 5.75           & 5.5                     \\ \cdashline{2-6} \noalign{\smallskip}
                                & mKMC                 & \textbf{1.25}  & \textbf{1.50}       & \textbf{1.25}  & \textbf{1.0}            \\ \bottomrule
\end{tabular}
\label{tab:sum_text_cluster}
\end{table}
Our approach can be utilized for any task where a human-understandable tabular representation can be extracted. We demonstrate this ability by extending our approach to perform text clustering on 4 text datasets: AG News\citep{agnewsyahooo}, DBPedia \citep{dbpedia}, 20 Newsgroups \citep{20news}, and Yahoo Answers \citep{agnewsyahooo}. For AG News, DBPedia, and Yahoo, we follow the approach taken in the previous literature \citep{subakti2022performance,wang2016semi} and sample 1,000 points from each class.

To create our interpretable representation, we remove the punctuation, lowercase, and tokenize all datapoints. We then lemmatize all tokens using the WordNet \citep{miller1995wordnet} lemmatizer available in the Natural Language Toolkit (NLTK) \citep{nltk} and remove english stopwords, corpus-specific stopwords that appear in more than 99\% of the datapoints, and rare words that appear in less than 1\% of the documents. Finally, we then calculate term-frequency in each datapoint and normalize the representation so that the $L_2$ norm of the resultant vector is equal to 1.0.

For the representation in our loss function (Equation (\ref{eqn:decouploss})), we leverage embeddings from the MPNet pretrained transformer \citep{song2020mpnet} available in the Sentence Transformers package \citep{sbert}. This model has a representation size of 768,  a maximum sequence length of 384, and uses mean pooling to construct its embedding.

To retain model sparsity, we opt to only use 128 single feature shape functions and 128 pairwise shape functions (when applicable). We set $m=1.025$ for our loss function (Equation (\ref{eqn:decouploss})).\footnote{On the text datasets, we empirically observed that $m=1.05$ resulted in degenerate solutions, therefore a smaller $m$ value was used.}

\subsubsection{Results}

Table \ref{tab:sum_text_cluster} provides a summary of the comparison between our approaches and our baselines on the text datasets. A detailed breakdown of the results by dataset can be found in Appendix \ref{sec:text_results}.

We observe that our approaches significantly outperform the interpretable baselines. Neur2CAM and NeurCAM obtain the first and second place, respectively, when considering the interpretable models across both external and internal metrics in both average value and average rank. More concretely, NeurCAM has a 167.9\% higher ARI than the SCT of depth seven and a 4.6\% lower Inertia than TAO with depth seven. The introduction of pairwise interactions makes this gap even more drastic, with Neur2CAM having both a higher average ARI (26. 6\% higher) and a lower average Inertia (1.5\% lower) than NeurCAM.

As expected, given the restrictions we imposed on our feature space, our interpretable approaches do not outperform the black-box model. However, despite significantly limiting the number of terms and interactions used,\footnote{We use a maximum of 128 interactions out more than 50,000 possible pairwise interactions.} we were able to obtain more than 92\% of the performance of the black-box model (Mini-Batch $K$-Means) across all external evaluation metrics. More specifically, Neur2CAM achieves an ARI equal to 92.7\% of the ARI, 95. 9\% of the NMI and 94. 8\% of the ACC achieved by Mini-Batch $K$-Means. Furthermore, Neur2CAM achieves an average Inertia value within 1.8\% of Mini-Batch $K$-Means. 

\subsection{Ablation Analysis}
\begin{table}[!ht]
\centering
\caption{Results of our ablation analysis. Full is our approach with both terms in the loss function. No CL denotes ablating the clustering loss term and No KL denotes ablating the KL-Divergence term.}
\label{tab:ablation}
\begin{tabular}{@{}cccccc@{}}
\toprule
                         &       & \multicolumn{2}{c}{NeurCAM}        & \multicolumn{2}{c}{Neur2CAM}       \\ 
                         &       & Loss      & Iner.            & 
 Loss      & Iner.            \\ \midrule
\multirow{3}{*}{Average} & Full  & \textbf{83.454} & \textbf{696.717} & \textbf{82.646} & \textbf{686.311} \\
                         & No CL & 84.254          & 698.432          & 83.012          & 686.824          \\
                         & No KL & 83.568          & 702.358          & 83.031          & 691.754          \\ \bottomrule
\end{tabular}
\end{table}
To demonstrate the benefit of the loss terms used in the second phase of our training, we introduce two ablations of NeurCAM. The first ablation removes the clustering loss (Equation (\ref{eqn:decouploss})) in the second phase of training, making NeurCAM to optimize only the KL divergence between NeurCAM and its relaxation. The second ablation keeps the clustering loss term, but instead ablates the KL-Divergence loss (Equation (\ref{eqn:kl})) so that NeurCAM only optimizes the clustering loss throughout its training. We perform this analysis on our text datasets and report both the loss (Equation (\ref{eqn:decouploss})) and the Inertia of the run that achieves the minimum Inertia across five seeds. We observe that including both terms in our loss function consistently results in lower values across both metrics. 

\subsection{Interpretability of NeurCAM}
In this section, we highlight the interpretability of NeurCAM and connect it back to the model-based interpretability desiderata presented in Section \ref{sec:modelbased}.
\subsubsection{Controllable Sparsity}
\begin{figure}[!ht]
    \centering
    \includegraphics[width = 0.6\columnwidth]{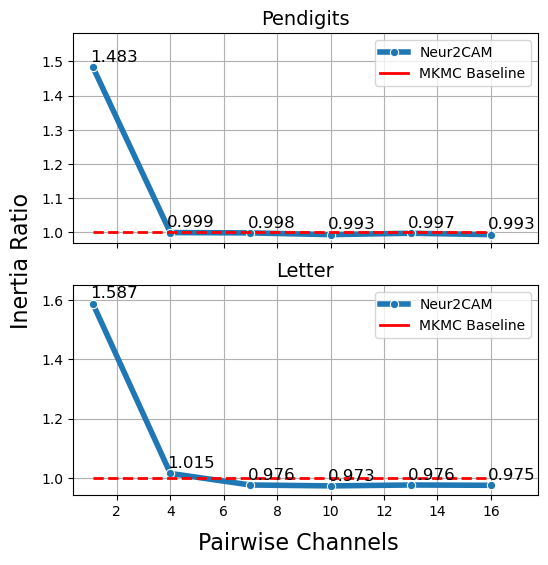}
    \caption{Cost Ratio of Neur2CAM compared to MKMC as more pairwise shape functions are added. }
    \label{fig:pendigitssensitivity}
\end{figure}
NeurCAM allows users to explicitly control the number of features and interactions used to construct the clusters through our selection gate mechanism (Equations (\ref{eqn:sel1}) and (\ref{eqn:sel2})). In many cases, users can significantly limit the number of shape functions used, improving the model sparsity while still generating high-quality clusters. To demonstrate this capability, we ablate the number of selection gates utilized. We report the ratio between the Inertia of Neur2CAM (minimum across five random seeds) and the Inertia of Mini-Batch $K$-Means (also across five random seeds) and plot the cost ratio as we vary the number of pairwise shape functions. To isolate the impact of using interactions, we set the number of single feature shape functions to zero for this experiment. In Figure \ref{fig:pendigitssensitivity}, we observe that we are able to achieve a loss value comparable to Mini-Batch $K$-Means with only 4 shape functions, or at most 4 pairwise interactions, when clustering on both the Pendigits and Letters dataset.

\subsubsection{Modularity and Simulability}
\begin{figure}[!ht]
    \centering
    \includegraphics[width = 0.6\columnwidth]{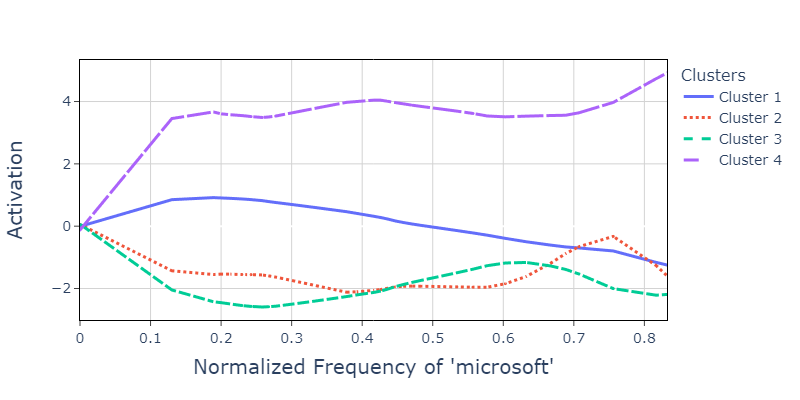}
    \caption{Shape Graphs for "Microsoft" learned when clustering the AG News Dataset}
    \label{fig:microsoft}
\end{figure}
The learned shape functions for each feature and interaction in NeurCAM are \emph{modular} and can be independently analyzed, allowing stakeholders to understand the impact individual features have on the end prediction as well as \emph{simulate} predictions by querying the different shape functions. As an example, Figure \ref{fig:microsoft} we displays the shape function learned by NeurCAM when clustering the AG News Text Dataset for the term ``Microsoft''. When examining this shape graph, we can see that when Microsoft is present in a sample, the network allocates more weight to Cluster 4 and reduces the weight to the other clusters. When examining the alignment between the clusters and the ground-truth labels, we observed that cluster 4 consists primarily of news titles related to technology, indicating that the presence of Microsoft in a sample results in NeurCAM mapping the sample to the ``technology'' cluster. A full example of NeurCAM's explanations can be found in Appendix \ref{sec:InterpretabilityExample}.

\section{Conclusion}

In this work, we present NeurCAM, the first approach to interpretable clustering that uses neural GAMs. Our experiments showcase our ability to produce high-quality clusters on tabular data that are comparable to black-box approaches while using a limited number of features and interactions chosen by our proposed selection gate mechanism. We also demonstrate NeurCAM's ability to utilize deep representations of the data while still providing explanations in a human interpretable feature space on both tabular and text datasets. 

NeurCAM is a powerful tool for large-scale clustering tasks for knowledge discovery and can be used as a foundation for further research on interpretable clustering via Generalized Additive Models. Potential extensions of our work involve the joint performance of clustering and representation learning by integrating approaches such as DEC / IEC \citep{xie2016unsupervised,guo2017improved} or VADE \citep{jiang2016variational}. Furthermore, other possible directions include designing specialized GAMs to focus on learned interpretable features from images, such as prototypes \citep{chen2019looks}, or time series, such as shapelets \citep{lines2012shapelet}. 

\bibliography{mybibfile}
\newpage
\appendix
\section{Official Implementation}
The official implementation for our approach can be accessed at \href{https://github.com/optimal-uoft/NeurCAM}{https://github.com/optimal-uoft/NeurCAM}. 
\section{Dataset Information}\label{sec:data_info_appendix}

\begin{table}[H]
    \begin{minipage}{.5\linewidth}
      \caption{Tabular Datasets}
      \centering
        \begin{tabular}{@{}llll@{}}
            \toprule
            Dataset   & $N$    & $D$ & $K$ \\ \midrule
            Adult      & 48,842  & 105 & 2 \\
            Avila & 20,867 &10 & 12 \\
            Gas Drift & 13,910 & 128 & 6 \\
            Letters & 20,000 & 16 & 26 \\
            Pendigits     & 10,992  & 16 & 10 \\
            Shuttle    & 58,000  & 9 & 7 \\\bottomrule
        \end{tabular}\label{tabl:tabular}
        
    \end{minipage}%
    \begin{minipage}{.5\linewidth}
      \centering
        \caption{Text Datasets}
        \begin{tabular}{@{}llll@{}}
        \toprule
            Dataset   & $N$    & $D$ & $K$ \\ \midrule
            AG News      & 4,000  & 338 & 4 \\
            DBPedia & 14,000 & 351 & 14 \\
            20 Newsgroups   & 18,846  & 1,065 & 20 \\
            Yahoo & 10,000 & 474 & 10 \\\bottomrule
        \end{tabular}\label{tabl:text}
    \end{minipage} 
\end{table}

\subsection{Tabular Representation Details}
Our SpectralNet \citep{spectralnet} configuration is as follows:
\begin{itemize}
    \item Spectral Net Hidden Dimensions: [512, 512, 2048, $K$]
    \item Siamese Net Hidden Dimensions: [512, 512, 2048, $K$]
\end{itemize}

All other parameters are the default presented by the original authors \citep{spectralnet}. 

Our DAE Configuration Configuration is as follows: 

\begin{itemize}
    \item Hidden Dimensions: [128, 64, 32, 16] (Reversed on for the decoder).
    \item Bottleneck Dimension: 8
    \item Input Dropout Probability: 0.1
\end{itemize}
The DAE is trained using the same configuration as NeurCAM.  

\subsubsection{Text Representation Details} \label{sec:text_repr_appendix}
To create our interpretable representation, we first remove the punctuation from all the datapoints and then tokenize then. We then lowercase all tokens and remove English stopwords. Then we lemmatize all words using the WordNet \citep{miller1995wordnet} lemmatizer available in the Natural Language Toolkit (NLTK) \citep{nltk}. Additionally, we remove corpus-specific stopwords that appear in more than 99\% of the datapoints and rare words that appear in less than 1\% of the documents. Finally we then calculate term frequency in each datapoint and normalize the representation such that the $L_2$ norm of the resultant vector is equal to 1.0.

For our contextual representation, we utilize embeddings from the MPNet transformer available in the Sentence Transformers package \citep{sbert}. This model has a representation size of 768,  a maximum sequence length of 384, and uses mean pooling to construct its embedding. More details about this model can be found at \href{https://huggingface.co/sentence-transformers/all-mpnet-base-v2}. 
\newpage

\section{Training and Implementation Details} \label{sec:hp}

NeurCAM is trained using Adam \citep{adam} for 1000 epochs and the learning rate is halved when the loss plateaus for 100 epochs. The selection gate tempering starts after 400 warmup epochs and continues for 100 epochs. For Neur2CAM, the pairwise gate temperature parameter $T_2$ is fully tempered before tempering $T$. To adjust for this additional tempering time, we allow Neur2CAM to also train for an extra 100 epochs as well.

To train the SCT, we follow \cite{SoftClusterTree} and opt to use the RMSProp optimizer for 1000 epochs with a cyclic scheduler. After the initial round of training, we slowly ramp the sparsity regularization weight by .001 every training step and terminate the training when each node of the tree utilizes only one feature \citep{SoftClusterTree}.

For Mini-Batch $K$-Means, we set the number of points used for initialization to five times the batch size and performed five initializations.

While the axis-aligned TAO described by \cite{gabidolla2022optimal} initializes their tree via ExKMC \citep{frost2020exkmc}, we instead initialize the tree via a CART decision tree fit on the results of $K$-Means \citep{kmeans} clustering as ExKMC does not support OSM on disentangled representations. TAO was run for 50 iterations.

\begin{table}[H]
\def\arraystretch{.9} 
\centering
\caption{Hyperparameter details for all approaches and baselines.}
\begin{tabular}{@{}cccccc@{}}
\toprule
                      & NeurCAM  & Neur2CAM & SCT      & TAO & Mkmc \\ \midrule
Batch Size            & 512      & 512      & 512      & --- & 512  \\
LR                    & 0.002    & 0.002    & 0.002    & --- & ---  \\
Optimizer             & Adam     & Adam     & RMSProp  & --- & ---  \\
LR Scheduler          & Plateau$^{(1)}$ & Plateau$^{(1)}$ & Cyclic$^{(2)}$ & --- & ---  \\
Hidden Dim.           & 256      & 256      & ---      & --- & ---  \\
N. Hidden             & 2        & 2        & ---      & --- & ---  \\
Activation            & ReLU     & ReLU     & ---      & --- & ---  \\
Warmup Steps          & 400      & 400      & ---      & --- & ---  \\
$T$ Tempering Steps     & 100      & 100      & ---      & --- & ---  \\
$T_2$ Tempering Steps    & ---      & 100      & ---      & --- & ---  \\
Total Epochs          & 1000     & 1100     & 1000     & 50  & 1000 \\
$B$              & 64       & 64       & ---      & --- & ---  \\
Branching Reg. Weight & ---      & ---      & 0.1      & --- & ---  \\
Sparsity Reg. Weight  & ---      & ---      & 0.001    & --- & ---  \\
$\gamma$                 & 1       & 1         & ---      & --- & ---  \\ 
$m$ & 1.05 or 1.025$^{(3)}$ & 1.05 or 1.025$^{(3)}$ & 1.05 or 1.025$^{(3)}$ & --- & ---\\
\bottomrule
\end{tabular}
\label{tab:hp}
\end{table}
(1) Halve learning rate when the loss does not improve for 100 epochs. (2) Cycled between 0.002 and 0.00002 every 100 epochs. (3) After the end of the main training epochs, this parameter is increased by 0.001 every training step until every node is fully sparse. For more information, we refer readers to \cite{SoftClusterTree}. (3) For tabular we use $m=1.05$, for text we use $m=1.025$. We empirically observed that a higher $m$ for text datasets resulted in a degenerate solution, therefore a smaller $m$ value was used. 

\newpage
\section{Full Tabular Results}\label{sec:tab_results}
\begin{table*}[!ht]
\def\arraystretch{.7} 
\centering
\small
\caption{Results from our tabular clustering experiments. The best overall results are highlighted in bold, while the best interpretable results are underlined. The dashed line separates interpretable and non-interpretable approaches.}
\begin{tabular}{@{}cccccccccc@{}}
\toprule
                           &       & \multicolumn{2}{c}{ARI}$\uparrow$               & \multicolumn{2}{c}{NMI}$\uparrow$      & \multicolumn{2}{c}{ACC}$\uparrow$                     & \multicolumn{2}{c}{Iner.}$\downarrow$                   \\ \midrule
                           &       & AE             & Spectral             & AE    & Spectral             & AE                   & Spectral             & AE                   & Spectral             \\ \midrule
\multirow{7}{*}{Adult}     & NCAM  & -0.018         & 0.000                & 0.002 & 0.000                & 0.720                & {\ul \textbf{0.761}} & {\ul 4.245}          & 0.994                \\
                           & N2CAM & -0.026         & 0.000                & 0.004 & 0.000                & 0.711                & {\ul \textbf{0.761}} & {\ul 4.245}          & 0.994                \\
                           & TAO-5 & -0.006         & 0.002                & 0.000 & 0.000                & 0.738                & 0.578                & 5.531                & 0.698                \\
                           & TAO-7 & -0.008         & -0.001               & 0.001 & 0.000                & 0.735                & 0.505                & 5.235                & {\ul 0.564}          \\
                           & SCT-5 & -0.007         & {\ul \textbf{0.043}} & 0.000 & 0.006                & 0.737                & 0.685                & 4.772                & 0.837                \\
                           & SCT-7 & -0.007         & {\ul \textbf{0.043}} & 0.000 & {\ul \textbf{0.009}} & 0.736                & 0.648                & 4.497                & 0.781                \\\cdashline{2-10} \noalign{\smallskip}
                           & mKMC  & -0.013         & -0.001               & 0.001 & 0.000                & 0.728                & 0.580                & \textbf{4.062}       & \textbf{0.121}       \\ \midrule
\multirow{7}{*}{Avila}     & NCAM  & 0.051          & 0.070                & 0.136 & 0.181                & 0.324                & 0.268                & 2.045                & 3.110                \\
                           & N2CAM & 0.069          & 0.072                & 0.143 & 0.176                & {\ul \textbf{0.338}} & 0.269                & 1.952                & {\ul 2.736}          \\
                           & TAO-5 & 0.063          & 0.046                & 0.153 & 0.135                & 0.335                & 0.258                & 2.450                & 5.538                \\
                           & TAO-7 & 0.056          & 0.067                & 0.133 & 0.165                & 0.327                & 0.273                & {\ul \textbf{1.281}} & 3.706                \\
                           & SCT-5 & 0.068          & {\ul \textbf{0.089}} & 0.131 & {\ul \textbf{0.193}} & 0.297                & 0.322                & 3.136                & 6.630                \\
                           & SCT-7 & 0.015          & 0.052                & 0.094 & 0.137                & 0.254                & 0.263                & 2.941                & 6.110                \\\cdashline{2-10} \noalign{\smallskip}
                           & mKMC  & 0.073          & 0.072                & 0.157 & 0.176                & 0.337                & 0.268                & 1.782                & \textbf{2.686}       \\ \midrule
\multirow{7}{*}{GasDrift}  & NCAM  & 0.068          & 0.120                & 0.247 & 0.259                & 0.363                & 0.407                & 2.502                & 1.617                \\
                           & N2CAM & 0.041          & {\ul \textbf{0.183}} & 0.199 & {\ul \textbf{0.326}} & 0.330                & {\ul \textbf{0.447}} & 3.216                & 2.493                \\
                           & TAO-5 & 0.065          & 0.162                & 0.241 & 0.266                & 0.357                & 0.427                & 2.187                & 1.852                \\
                           & TAO-7 & 0.062          & 0.164                & 0.236 & 0.278                & 0.355                & 0.430                & {\ul \textbf{1.877}} & 1.552                \\
                           & SCT-5 & 0.124          & 0.069                & 0.255 & 0.198                & 0.391                & 0.332                & 3.462                & 2.169                \\
                           & SCT-7 & 0.109          & 0.078                & 0.235 & 0.197                & 0.393                & 0.365                & 3.346                & 1.977                \\\cdashline{2-10} \noalign{\smallskip}
                           & mKMC  & 0.100          & 0.167                & 0.226 & 0.285                & 0.355                & 0.432                & 2.858                & \textbf{1.477}       \\ \midrule
\multirow{7}{*}{Letter}    & NCAM  & 0.151          & {\ul 0.242}          & 0.373 & {\ul 0.545}          & 0.260                & 0.402                & 1.844                & 5.154                \\
                           & N2CAM & 0.173          & {\ul 0.242}          & 0.402 & 0.540                & 0.299                & {\ul 0.406}          & {\ul 1.624}          & {\ul 4.702}          \\
                           & TAO-5 & 0.115          & 0.161                & 0.343 & 0.418                & 0.241                & 0.331                & 2.612                & 15.332               \\
                           & TAO-7 & 0.139          & 0.192                & 0.375 & 0.475                & 0.271                & 0.382                & 1.953                & 8.104                \\
                           & SCT-5 & 0.139          & 0.171                & 0.344 & 0.383                & 0.240                & 0.281                & 3.114                & 18.116               \\
                           & SCT-7 & 0.128          & 0.146                & 0.326 & 0.384                & 0.242                & 0.294                & 3.046                & 15.125               \\\cdashline{2-10} \noalign{\smallskip}
                           & mKMC  & 0.156          & \textbf{0.256}       & 0.384 & \textbf{0.551}       & 0.286                & \textbf{0.418}       & \textbf{1.403}       & \textbf{3.697}       \\ \midrule
\multirow{7}{*}{Pendigits} & NCAM  & 0.506          & 0.754                & 0.629 & 0.822                & 0.698                & 0.871                & {\ul 1.344}          & 0.712                \\
                           & N2CAM & 0.502          & {\ul \textbf{0.757}} & 0.643 & {\ul 0.826}          & 0.653                & {\ul \textbf{0.872}} & 1.355                & {\ul 0.648}          \\
                           & TAO-5 & 0.481          & 0.692                & 0.607 & 0.770                & 0.681                & 0.841                & 1.610                & 1.196                \\
                           & TAO-7 & 0.493          & 0.743                & 0.626 & 0.817                & 0.693                & 0.867                & 1.406                & 0.762                \\
                           & SCT-5 & 0.402          & 0.520                & 0.548 & 0.636                & 0.550                & 0.730                & 1.954                & 3.052                \\
                           & SCT-7 & 0.458          & 0.670                & 0.597 & 0.732                & 0.646                & 0.824                & 1.790                & 1.716                \\\cdashline{2-10} \noalign{\smallskip}
                           & mKMC  & 0.489          & 0.754                & 0.624 & \textbf{0.831}       & 0.677                & \textbf{0.872}       & \textbf{1.330}       & \textbf{0.611}       \\ \midrule
\multirow{7}{*}{Shuttle}   & NCAM  & 0.392          & 0.380                & 0.428 & {\ul \textbf{0.500}} & {\ul \textbf{0.778}} & 0.654                & 3.072                & {\ul \textbf{0.383}} \\
                           & N2CAM & 0.382          & 0.376                & 0.377 & 0.494                & 0.771                & 0.653                & {\ul \textbf{2.002}} & 0.390                \\
                           & TAO-5 & 0.393          & 0.376                & 0.378 & 0.493                & 0.775                & 0.653                & 3.367                & 0.470                \\
                           & TAO-7 & {\ul 0.394}    & 0.375                & 0.377 & 0.493                & 0.775                & 0.652                & 2.135                & 0.412                \\
                           & SCT-5 & 0.230          & 0.245                & 0.376 & 0.465                & 0.494                & 0.521                & 4.871                & 1.955                \\
                           & SCT-7 & 0.239          & 0.199                & 0.349 & 0.429                & 0.544                & 0.461                & 4.875                & 0.946                \\\cdashline{2-10} \noalign{\smallskip}
                           & mKMC  & \textbf{0.460} & 0.249                & 0.411 & 0.435                & 0.691                & 0.549                & 4.655                & 0.881                \\ \midrule
\multirow{7}{*}{Average}   & NCAM  & 0.192          & 0.261                & 0.303 & 0.385                & 0.524                & 0.561                & 2.509                & 1.995                \\
                           & N2CAM & 0.190          & {\ul \textbf{0.272}} & 0.295 & {\ul \textbf{0.394}} & 0.517                & {\ul \textbf{0.568}} & 2.399                & {\ul 1.994}          \\
                           & TAO-5 & 0.185          & 0.240                & 0.287 & 0.347                & 0.521                & 0.515                & 2.960                & 4.181                \\
                           & TAO-7 & 0.189          & 0.257                & 0.291 & 0.371                & 0.526                & 0.518                & {\ul 2.315}          & 2.517                \\
                           & SCT-5 & 0.159          & 0.190                & 0.276 & 0.314                & 0.452                & 0.479                & 3.552                & 5.460                \\
                           & SCT-7 & 0.157          & 0.198                & 0.267 & 0.315                & 0.469                & 0.476                & 3.416                & 4.443                \\\cdashline{2-10} \noalign{\smallskip}
                           & mKMC  & 0.211          & 0.250                & 0.301 & 0.380                & 0.512                & 0.520                & \textbf{2.682}       & \textbf{1.579}       \\ \bottomrule
\end{tabular}
\label{tab:tab_cluster}
\end{table*} 
\newpage
\section{Full Text Results} \label{sec:text_results}
\begin{table}[H]
\centering
\small
\caption{Results from our text clustering experiments. The best overall results are highlighted in bold, while the best interpretable results are underlined. The dashed line separates interpretable and non-interpretable approaches.}
\begin{tabular}{@{}cccccc@{}}
\toprule
                            &       & ARI $\uparrow$                 & NMI $\uparrow$                 & ACC $\uparrow$                 & Iner. $\downarrow$           \\ \midrule
\multirow{7}{*}{AG News}    & NCAM  & 0.464                & 0.462                & 0.755                & 728.099          \\
                            & N2CAM & {\ul \textbf{0.609}} & {\ul \textbf{0.576}} & {\ul \textbf{0.828}} & {\ul 720.732}    \\
                            & TAO-5 & 0.035                & 0.163                & 0.405                & 751.176          \\
                            & TAO-7 & 0.075                & 0.181                & 0.471                & 750.871          \\
                            & SCT-5 & 0.209                & 0.288                & 0.562                & 742.914          \\
                            & SCT-7 & 0.252                & 0.282                & 0.563                & 742.650          \\\cdashline{2-6} \noalign{\smallskip}
                            & mKMC  & 0.538                & 0.548                & 0.762                & \textbf{719.803} \\ \midrule
\multirow{7}{*}{DBPedia}    & NCAM  & 0.636                & 0.761                & {\ul 0.770}          & 640.994          \\
                            & N2CAM & {\ul 0.676}          & {\ul 0.802}          & 0.738                & {\ul 638.379}    \\
                            & TAO-5 & 0.152                & 0.512                & 0.375                & 697.140          \\
                            & TAO-7 & 0.218                & 0.561                & 0.471                & 681.172          \\
                            & SCT-5 & 0.286                & 0.564                & 0.514                & 690.059          \\
                            & SCT-7 & 0.262                & 0.546                & 0.452                & 688.185          \\\cdashline{2-6} \noalign{\smallskip}
                            & mKMC  & \textbf{0.774}       & \textbf{0.842}       & \textbf{0.845}       & \textbf{627.486} \\ \midrule
\multirow{7}{*}{Newsgroups} & NCAM  & 0.185                & 0.388                & 0.404                & 682.520          \\
                            & N2CAM & {\ul 0.296}          & {\ul 0.477}          & {\ul 0.471}          & {\ul 661.685}    \\
                            & TAO-5 & 0.012                & 0.169                & 0.161                & 744.979          \\
                            & TAO-7 & 0.013                & 0.215                & 0.203                & 735.034          \\
                            & SCT-5 & 0.006                & 0.053                & 0.099                & 759.167          \\
                            & SCT-7 & 0.013                & 0.051                & 0.121                & 756.019          \\\cdashline{2-6} \noalign{\smallskip}
                            & mKMC  & \textbf{0.400}       & \textbf{0.561}       & \textbf{0.550}       & \textbf{630.874} \\ \midrule
\multirow{7}{*}{Yahoo}      & NCAM  & 0.150                & 0.240                & 0.407                & 735.257          \\
                            & N2CAM & {\ul 0.240}          & {\ul \textbf{0.329}} & {\ul 0.465}          & {\ul 724.446}    \\
                            & TAO-5 & 0.012                & 0.099                & 0.201                & 756.741          \\
                            & TAO-7 & 0.018                & 0.132                & 0.239                & 754.356          \\
                            & SCT-5 & 0.009                & 0.027                & 0.156                & 762.317          \\
                            & SCT-7 & 0.008                & 0.019                & 0.143                & 764.162          \\\cdashline{2-6} \noalign{\smallskip}
                            & mKMC  & \textbf{0.251}       & 0.328                & \textbf{0.480}       & \textbf{718.121} \\ \midrule
\multirow{7}{*}{Average}    & NCAM  & 0.359                & 0.463                & 0.584                & 696.717          \\
                            & N2CAM & {\ul 0.455}          & {\ul 0.546}          & {\ul 0.625}          & {\ul 686.311}    \\
                            & TAO-5 & 0.052                & 0.236                & 0.286                & 737.509          \\
                            & TAO-7 & 0.081                & 0.272                & 0.346                & 730.358          \\
                            & SCT-5 & 0.127                & 0.233                & 0.333                & 738.614          \\
                            & SCT-7 & 0.134                & 0.224                & 0.320                & 737.754          \\\cdashline{2-6} \noalign{\smallskip}
                            & mKMC  & \textbf{0.491}       & \textbf{0.570}       & \textbf{0.659}       & \textbf{674.071} \\ \bottomrule
\end{tabular}
\label{tab:text_cluster}
\end{table}
\newpage

\section{NeurCAM Explanation Consistency}
NeurCAM allows users to limit the number of features used to build clusters by setting the number of shape functions NeurCAM learns to a small number. To evaluate the consistency of the selection mechanism across different runs of the algorithm, we report the selected features shared across different runs of NeurCAM on our text datasets. As NeurCAM may converge to a locally optimum solution, we perform multiple runs and select the runs with the lowest Inertia values for the hard clustering, similar to the evaluation setting of our tabular and text clustering experiment. For this set of experiments, we run fifteen random seeds and choose the five runs with the lowest Inertia values for the hard clustering. We then select the top $K$ features according to importance. Finally, we calculate the average intersection of the top $K$ words between each pair of runs for different $K$ values. The results of this analysis can be seen in Table \ref{tab:Intersect}. In addition, we provide the top ten important words for the runs for each dataset in Tables \ref{tab:AG_News}, \ref{tab:DBPedia}, \ref{tab:20Newsgroups}, and \ref{tab:Yahoo}. We observe that the runs share a significant intersection in the top-$K$ features with more than a 50\% average intersection across all the various $K$ values. 

\begin{table}[H]
\centering
\def\arraystretch{.9}
\caption{Average number of shared important terms across pairs of runs.}
\label{tab:Intersect}
\begin{tabular}{@{}cccccc@{}}
\toprule
           & \multicolumn{5}{c}{Average Intersection Size  @ $K$} \\ \midrule
Dataset    & $K=$ 10     & $K=$ 25      & $K=$ 50      & $K=$ 75      & $K=$ 100    \\ \midrule
AG News    & 5.3    & 16.8    & 29.7    & 43.3    & 55.6   \\
DBPedia    & 8.4    & 19.3    & 36.8    & 49.5    & 62.4   \\
Newsgroups & 5.5    & 14.9    & 29.1    & 40.9    & 55.1   \\
Yahoo      & 5.9    & 14.3    & 32.5    & 45.7    & 55.8   \\ \bottomrule
\end{tabular}
\end{table}
\begin{table}[H]
\centering
\begin{tabular}{@{}lllll@{}}
\toprule
\multicolumn{5}{c}{AG News}                                \\ \midrule
company   & iraq      & company   & company   & company    \\
iraq      & company   & 39s       & iraq      & price      \\
corp      & corp      & iraq      & microsoft & corp       \\
game      & microsoft & computer  & internet  & technology \\
player    & internet  & microsoft & market    & game       \\
new       & oil       & game      & oil       & iraq       \\
market    & team      & oil       & team      & night      \\
microsoft & new       & software  & player    & microsoft  \\
software  & price     & corp      & victory   & internet   \\
sunday    & season    & market    & president & billion    \\ \bottomrule
\end{tabular}
\caption{Top 10 Important words for five different runs of NeurCAM for the AG News dataset.}
\label{tab:AG_News}
\end{table}

\begin{table}[H]
\centering
\begin{tabular}{@{}lllll@{}}
\toprule
\multicolumn{5}{c}{DBPedia}                          \\ \midrule
born     & born     & born     & born     & specie   \\
located  & specie   & located  & specie   & born     \\
built    & located  & film     & film     & film     \\
specie   & genus    & built    & located  & located  \\
film     & district & school   & county   & district \\
album    & family   & genus    & village  & village  \\
village  & film     & district & school   & school   \\
district & village  & specie   & built    & family   \\
released & school   & family   & district & genus    \\
school   & built    & released & album    & built    \\ \bottomrule
\end{tabular}
\caption{Top 10 Important words for five different runs of NeurCAM for the DBPedia dataset.}
\label{tab:DBPedia}
\end{table}
\begin{table}[H]
\centering
\begin{tabular}{@{}lllll@{}}
\toprule
\multicolumn{5}{c}{Newsgroups}                                 \\ \midrule
game       & window     & window     & window     & people     \\
work       & government & car        & know       & window     \\
team       & people     & people     & team       & car        \\
government & car        & game       & car        & god        \\
car        & god        & program    & card       & christian  \\
year       & program    & government & government & government \\
im         & problem    & software   & drive      & drive      \\
window     & software   & god        & christian  & game       \\
problem    & card       & thanks     & driver     & price      \\
drive      & game       & team       & space      & software   \\ \bottomrule
\end{tabular}
\caption{Top 10 Important words for five different runs of NeurCAM for the 20Newsgroups dataset.}
\label{tab:20Newsgroups}
\end{table}

\begin{table}[H]
\centering
\begin{tabular}{@{}lllll@{}}
\toprule
\multicolumn{5}{c}{Yahoo}                         \\ \midrule
help     & help   & help     & think    & think   \\
think    & people & people   & guy      & girl    \\
song     & think  & god      & help     & people  \\
people   & girl   & girl     & girl     & god     \\
girl     & god    & guy      & dont     & guy     \\
guy      & guy    & want     & people   & dont    \\
love     & want   & computer & world    & love    \\
god      & world  & problem  & woman    & year    \\
know     & love   & jesus    & computer & problem \\
computer & know   & religion & american & friend  \\ \bottomrule
\end{tabular}
\caption{Top 10 Important words for five different runs of NeurCAM for the Yahoo dataset.}
\label{tab:Yahoo}
\end{table}
\section{NeurCAM Time Analysis}
In Table \ref{tab:time}, we provide the runtime of NeurCAM and Neur2CAM on the text dataset. This set of experiments was run on a Nvidia GEFORCE 3080 RTX with 10GB VRAM, 32GB RAM, and a 12th Gen Intel(R) Core(TM) i7-12700 @  2.10 GHz. All results are across five seeds. Note that we do not report the runtime of TAO as this approach is not trained using scalable mini-batch methods, making the performance of TAO incomparable to the runtime of NeurCAM and the SCT. Additionally, Minbatch $K$-Means (a uninterpretable clustering model) is omitted from the table as it is used to initialize the centroids for our model, making the initialization time of NeurCAM equivalent to the time required to run Minibatch $K$-Means.




\begin{table}[H]
\centering
\def\arraystretch{.9}
\caption{Average Training time in seconds.}
\label{tab:time}
\begin{tabular}{@{}ccccc@{}}
\toprule
                            &       & Initialization & Warmup  & Total   \\ \midrule
\multirow{3}{*}{AGNews}     & NCAM  & 0.05           & 168.72  & 426.45  \\
                            & N2CAM & 0.06           & 312.26  & 873.13  \\
                            & SCT-7   & ---            & ---     & 1001.45 \\ \midrule
\multirow{3}{*}{DBPedia}    & NCAM  & 0.19           & 515.85  & 1308.47 \\
                            & N2CAM & 0.19           & 961.75  & 2674.51 \\
                            & SCT-7   & ---            & ---     & 2353.12 \\ \midrule
\multirow{3}{*}{Newsgroups} & NCAM  & 0.30           & 664.77  & 1696.48 \\
                            & N2CAM & 0.30           & 1261.56 & 3508.65 \\
                            & SCT-7   & ---            & ---     & 3032.01 \\ \midrule
\multirow{3}{*}{Yahoo}      & NCAM  & 0.10           & 368.99  & 935.36  \\
                            & N2CAM & 0.10           & 708.54  & 1966.50 \\
                            & SCT-7   & ---            & ---     & 1401.55 \\ \bottomrule
\end{tabular}
\end{table}

\begin{table}[H]
\centering
\def\arraystretch{.95}
\caption{Standard Deviation of Training Time in seconds. }
\label{tab:my-table}
\begin{tabular}{@{}ccccc@{}}
\toprule
                            &       & Initialization & Warmup & Total  \\ \midrule
\multirow{3}{*}{AGNews}     & NCAM  & 0.02           & 5.19   & 2.83   \\
                            & N2CAM & 0.01           & 3.93   & 6.73   \\
                            & SCT   & ---            & ---    & 119.15 \\ \midrule
\multirow{3}{*}{DBPedia}    & NCAM  & 0.02           & 5.17   & 3.60   \\
                            & N2CAM & 0.03           & 3.69   & 12.00  \\
                            & SCT   & ---            & ---    & 32.70  \\ \midrule
\multirow{3}{*}{Newsgroups} & NCAM  & 0.03           & 10.41  & 13.38  \\
                            & N2CAM & 0.03           & 7.87   & 21.77  \\
                            & SCT   & ---            & ---    & 21.65  \\ \midrule
\multirow{3}{*}{Yahoo}      & NCAM  & 0.01           & 2.72   & 6.68   \\
                            & N2CAM & 0.01           & 4.42   & 5.67   \\
                            & SCT   & ---            & ---    & 13.51  \\ \bottomrule
\end{tabular}
\end{table}

\section{Interpretability Example} \label{sec:InterpretabilityExample}
The interpretability of NeurCAM comes from the fact that the learned shape functions for each feature and interaction can be visualized and summarized, allowing stakeholders to understand the impact individual features have on the end prediction. In this section, we demonstrate this capability on the Shuttle dataset and provide visualizations of the explanations from NeurCAM.  The Shuttle dataset contains 9 numerical attributes and the full information on the dataset can be found at the following link: \href{https://archive.ics.uci.edu/dataset/148/statlog+shuttle}{https://archive.ics.uci.edu/dataset/148/statlog+shuttle}.

For this, set of experiments, we allow for nine single-feature selection shape functions and one pairwise gate. Our interpretable representation is the original feature space and the representation used by the loss function are deep embeddings from a pre-trained Denoising AutoEncoder. This model achieved an ARI of 0.485, an NMI of 0.424, an ACC of 0.738, and an Inertia of 2.88.

\begin{figure}[H]
     \centering
     \begin{subfigure}[b]{0.45\textwidth}
        \centering
        \includegraphics[width=\textwidth]{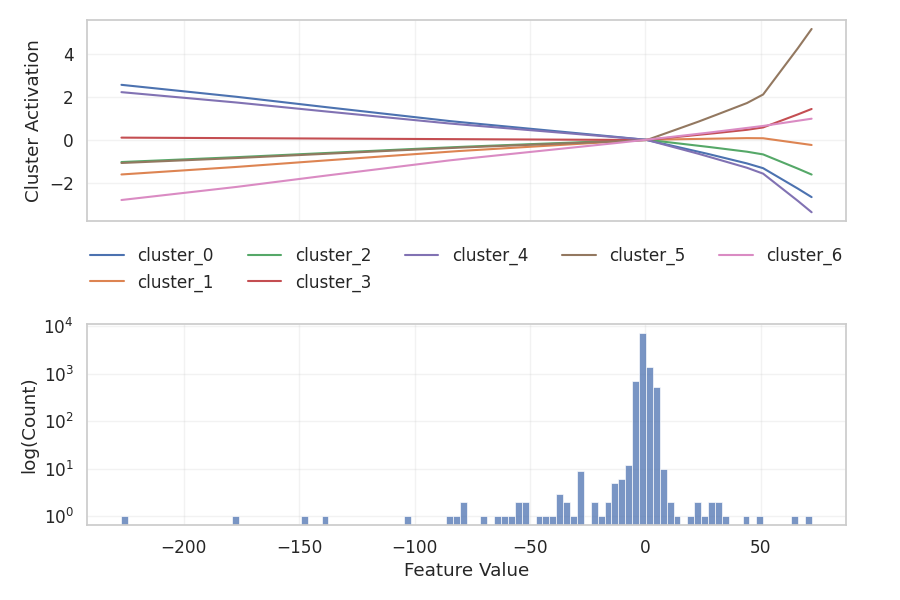}
        \caption{Feature A2}
        \label{fig:a2shape}
     \end{subfigure}
     \hfill
     \begin{subfigure}[b]{0.45\textwidth}
        \centering
        \includegraphics[width=\linewidth]{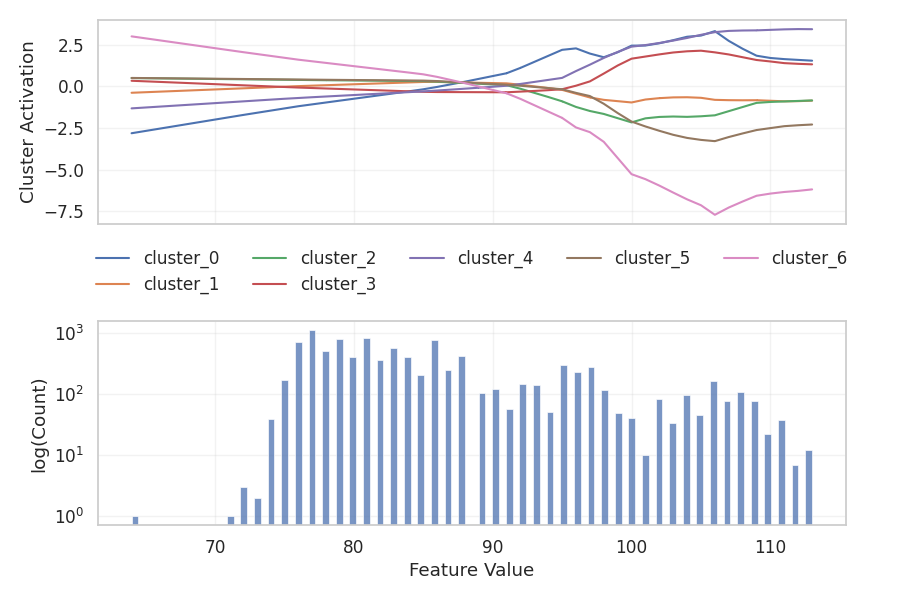}
        \caption{Feature A3}
        \label{fig:a3shape}
     \end{subfigure} \\
     \begin{subfigure}[b]{0.45\textwidth}
        \centering
        \includegraphics[width=\textwidth]{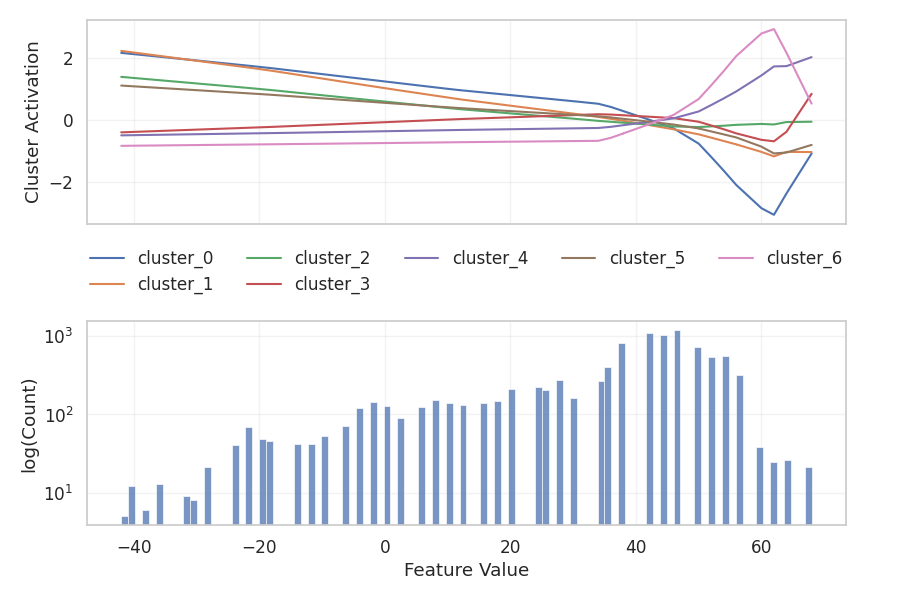}
        \caption{Feature A5}
        \label{fig:a5shape}
     \end{subfigure}
     \hfill
     \begin{subfigure}[b]{0.45\textwidth}
        \centering
        \includegraphics[width=\linewidth]{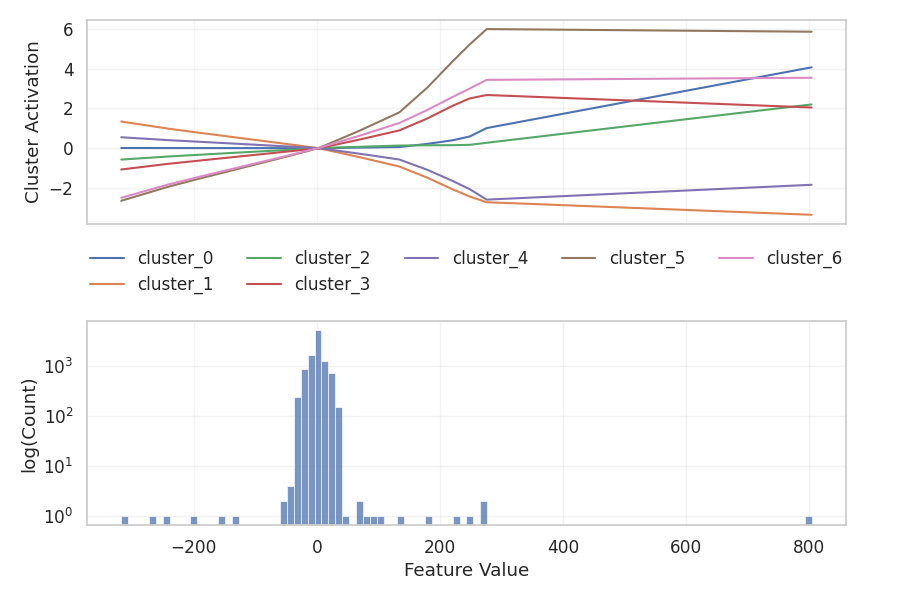}
        \caption{Feature A6}
        \label{fig:a6shape}
     \end{subfigure} \\
     \begin{subfigure}[b]{0.45\textwidth}
        \centering
        \includegraphics[width=\textwidth]{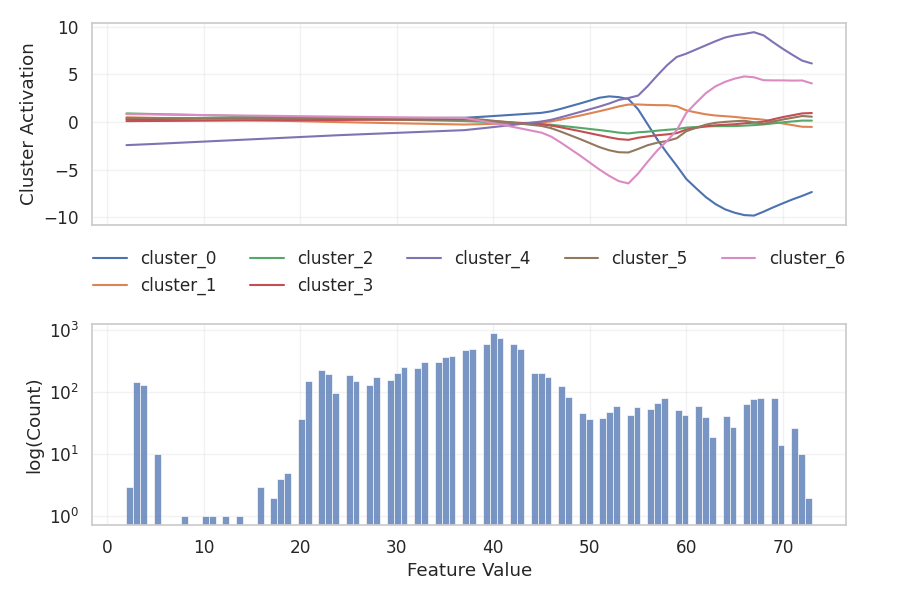}
        \caption{Feature A7}
        \label{fig:a7shape}
     \end{subfigure}
     \hfill
     \begin{subfigure}[b]{0.45\textwidth}
        \centering
        \includegraphics[width=\linewidth]{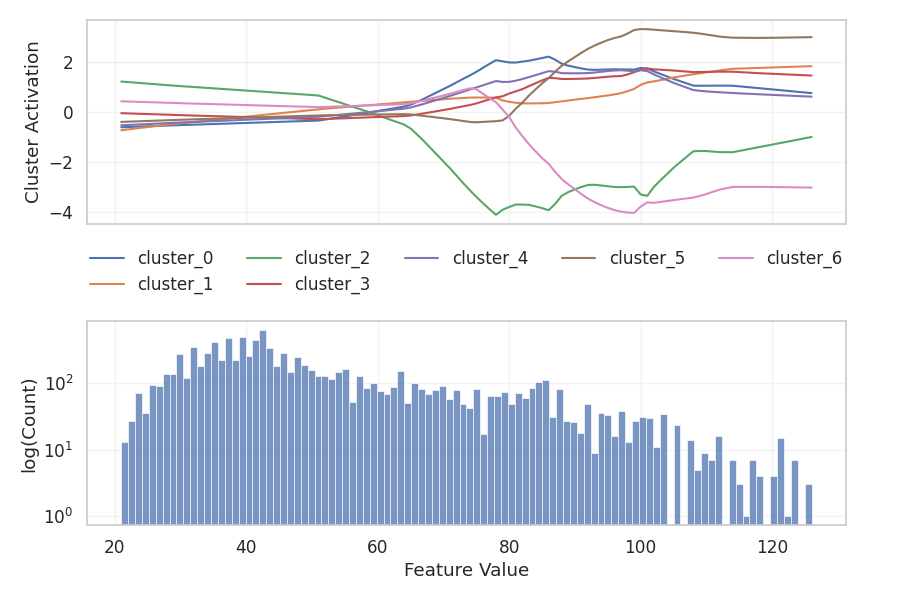}
        \caption{Feature A8}
        \label{fig:a8hape}
     \end{subfigure} \\
     \begin{subfigure}[b]{0.45\textwidth}
        \centering
        \includegraphics[width=\textwidth]{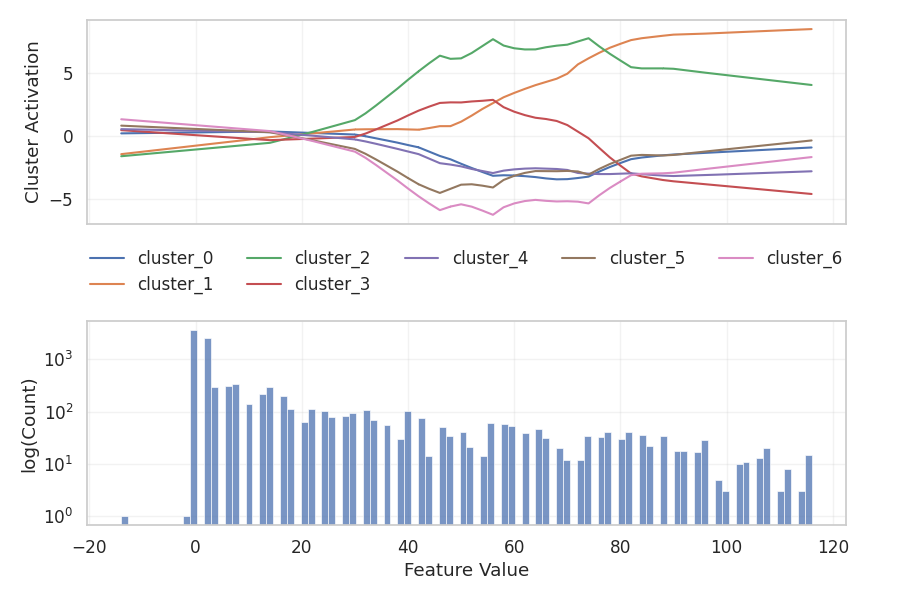}
        \caption{Feature A9}
        \label{fig:a9shape}
     \end{subfigure}

     \caption{Single Feature Shape Graphs for the Shuttle Dataset (Top) as well as feature value histograms (bottom)}
    \label{fig:o1ShapeGraphs}
\end{figure}

\paragraph{Single Feature Shape Graphs:} We first plot the single-feature shape graphs for NeurCAM along with the distribution of values of the features (Figure \ref{fig:o1ShapeGraphs}). Note that while we allowed at most nine features, NeurCAM only selected seven features. Examining these shape graphs, we observe some interesting behaviors. For example, we observe that feature A3 (Figure \ref{fig:a3shape}) seems to be important in determining whether or not a point is in cluster 6, as when the value of A3 is high, NeurCAM reduces the weight assigned to cluster 6, but increases it slightly when the value of A3 is low. Another observed pattern is that when the value of feature A7 (Figure \ref{fig:a7shape}) is high (above 60), NeurCAM assigns a high weight to cluster 4 and 6, but significantly reduces the weight to cluster 0. 

\begin{figure}[H]
     \centering
     \begin{subfigure}[b]{0.45\textwidth}
        \centering
        \includegraphics[width=\textwidth]{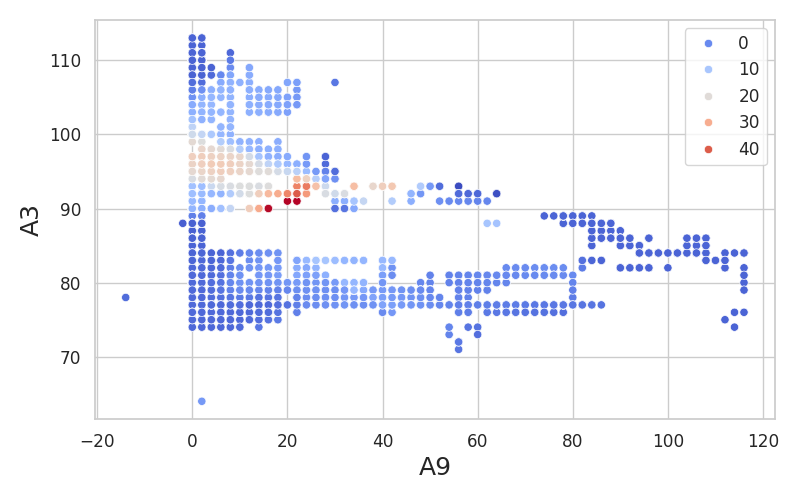}
        \caption{Cluster 0}
        \label{fig:c0pair}
     \end{subfigure}
     \hfill
     \begin{subfigure}[b]{0.45\textwidth}
        \centering
        \includegraphics[width=\linewidth]{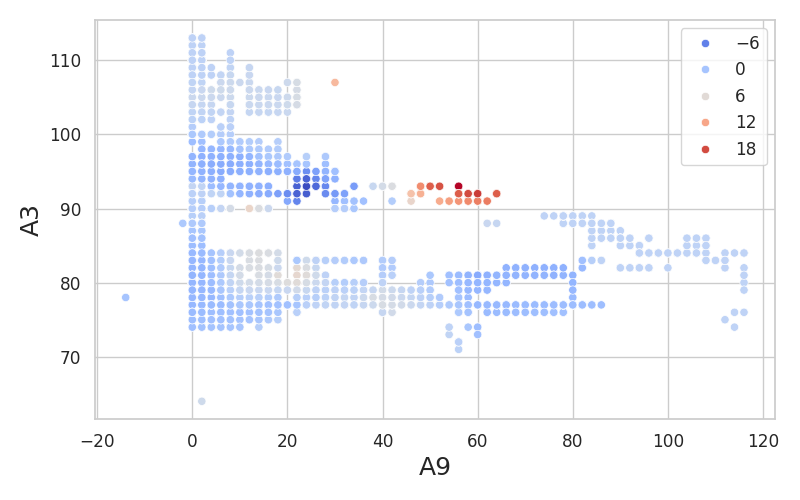}
        \caption{Cluster 1}
        \label{fig:c1pair}
     \end{subfigure} \\
     \begin{subfigure}[b]{0.45\textwidth}
        \centering
        \includegraphics[width=\textwidth]{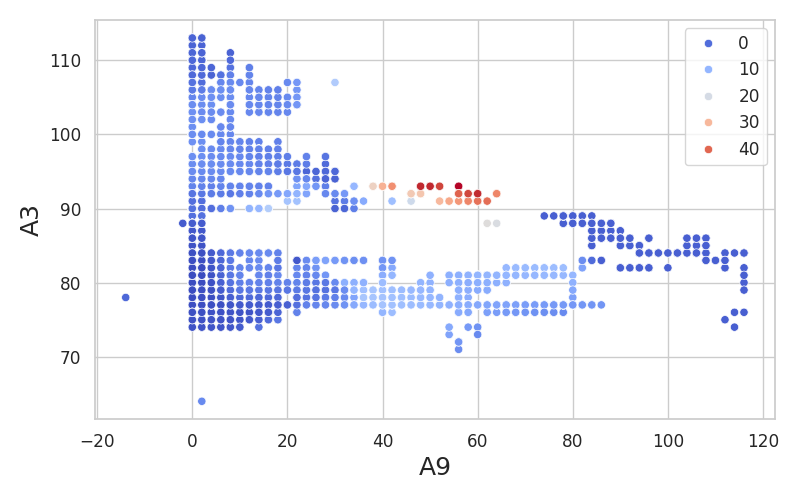}
        \caption{Cluster 2}
        \label{fig:c2pair}
     \end{subfigure}
     \hfill
     \begin{subfigure}[b]{0.45\textwidth}
        \centering
        \includegraphics[width=\linewidth]{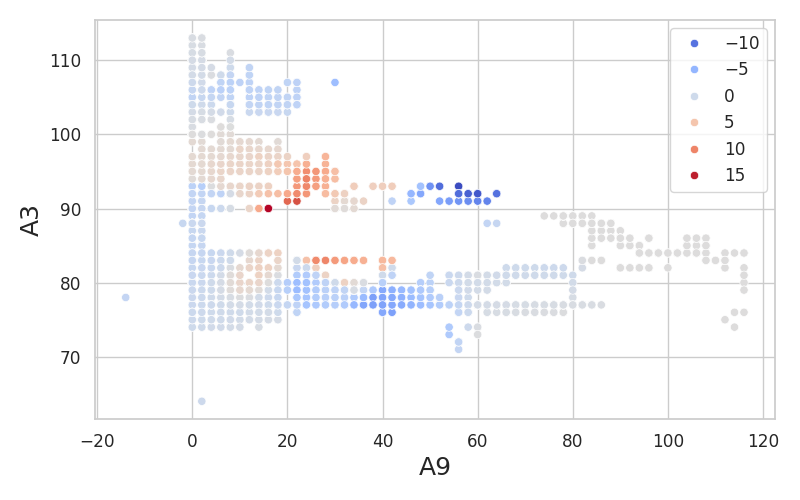}
        \caption{Cluster 3}
        \label{fig:c3pair}
     \end{subfigure} \\
     \begin{subfigure}[b]{0.45\textwidth}
        \centering
        \includegraphics[width=\textwidth]{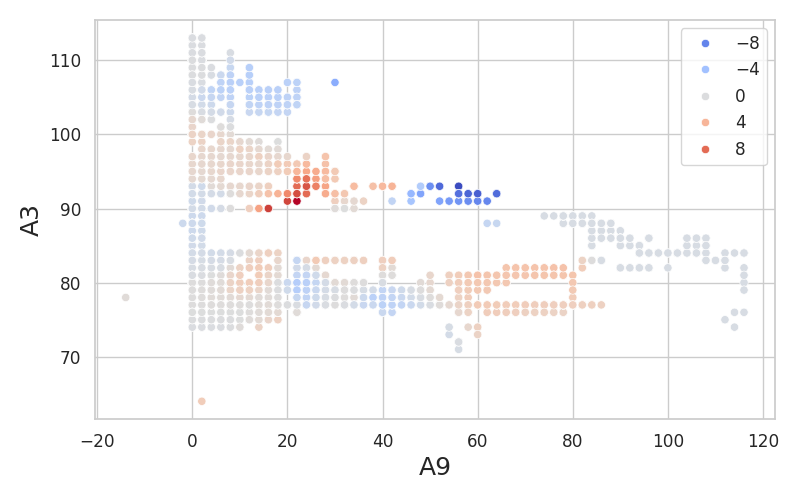}
        \caption{Cluster 4}
        \label{fig:c4pair}
     \end{subfigure}
     \hfill
     \begin{subfigure}[b]{0.45\textwidth}
        \centering
        \includegraphics[width=\linewidth]{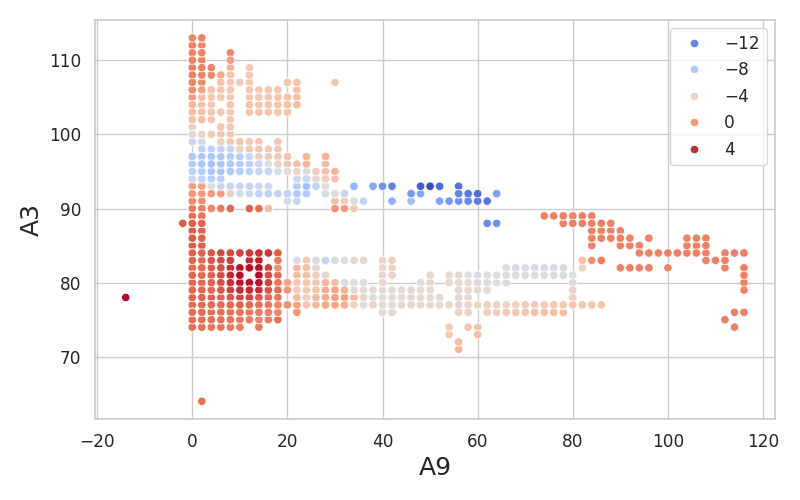}
        \caption{Cluster 5}
        \label{fig:c5pair}
     \end{subfigure} \\
     \begin{subfigure}[b]{0.45\textwidth}
        \centering
        \includegraphics[width=\textwidth]{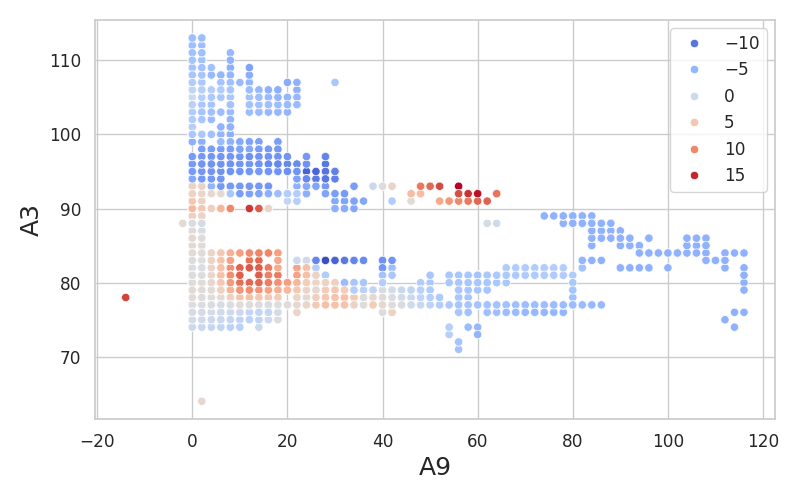}
        \caption{Cluster 6}
        \label{fig:c6pair}
     \end{subfigure}
     
     \caption{Pairwise Feature Shape Graph for the Shuttle Dataset}
    \label{fig:o2ShapeGraphs}
\end{figure}

\paragraph{Pairwise Shape Graphs: } The pairwise interactions of a GA$^2$M can be visualized as a set of heatmaps, and we plot the pairwise shape graph learned by NeurCAM in Figure \ref{fig:o2ShapeGraphs} following the format utilized by \cite{NODEGAM}. Examining these shape graphs, we can see some interesting patterns. For example, when A3 is approximately equal to 90 and A9 is approximately equal to 20, NeurCAM assigns more weight to clusters 0 (Figure \ref{fig:c0pair}), 3 (Figure \ref{fig:c3pair}), and 4 (Figure \ref{fig:c4pair}) while reducing the weights assigned to cluster 5 (Figure \ref{fig:c5pair}). 
\newpage

\end{document}